%% file: top.tex
\documentclass[10pt,twocolumn,letterpaper]{article}

\usepackage{cvpr}
\usepackage{times}
\usepackage{epsfig}
\usepackage{graphicx}
\usepackage{amsmath}
\usepackage{amssymb}

\usepackage[pagebackref=true,breaklinks=true,letterpaper=true,colorlinks,bookmarks=false]{hyperref}

\cvprfinalcopy % *** Uncomment this line for the final submission

\def\cvprPaperID{4159} % *** Enter the CVPR Paper ID here

\ifcvprfinal\fi
\begin{document}

%%%%%%%%% TITLE
\title{Neural Scene Decomposition for \NEW{Multi-Person} Motion Capture}
% OLD: Neural Scene Decomposition for Human Motion Capture

\author{Helge Rhodin \quad Victor  Constantin \quad Isinsu  Katircioglu\quad Mathieu  Salzmann\quad Pascal  Fua \\
	CVLab, EPFL,
	Lausanne, Switzerland\\
	{helge.rhodin@epfl.ch}
}

%\author{Helge  Rhodin \quad Victor  Constantin\and Isinsu  Katircioglu\and Mathieu  Salzmann\and Pascal  Fua
%	CVLab, EPFL, Lausanne, Switzerland
%}

%\authorrunning{Helge  Rhodin \and Mathieu Salzmann \and Pascal  Fua}

\input{tex/defs.tex}

\maketitle
%\thispagestyle{empty}

%%%%%%%%% ABSTRACT

%%%%%%%%% BODY TEXT

\input{tex/abstract.tex}
\input{tex/introduction.tex}
\input{tex/related.tex}
\input{tex/method.tex}
\input{tex/evaluation.tex}
\input{tex/conc.tex}

\parag{Acknowledgments.}\hspace{-0.2cm}This work was supported by the Swiss National Science Foundation and a Microsoft JRC Project.

\clearpage
\section*{Appendix}
\appendix
\input{tex/appendix.tex}

\clearpage
{\small
	\bibliographystyle{ieee}
	\bibliography{string,vision,graphics,learning,biomed,misc}
}

\end{document}

%% file: tex/defs.tex
% !TEX root = ../top.tex
% !TEX spellcheck = en-US

%\definecolor{olive}{RGB}{50,150,50}

\newif\ifdraft
%\drafttrue

\ifdraft
\newcommand{\PF}[1]{{\color{red}{\bf pf: #1}}}
\newcommand{\pf}[1]{{\color{red} #1}}
\newcommand{\HR}[1]{{\color{blue}{\bf hr: #1}}}
\newcommand{\hr}[1]{{\color{blue} #1}}
\newcommand{\VC}[1]{{\color{green}{\bf vc: #1}}}
\newcommand{\vc}[1]{{\color{green} #1}}
\newcommand{\ms}[1]{{\color{magenta}{#1}}}
\newcommand{\MS}[1]{{\color{magenta}{\bf ms: #1}}}
\newcommand{\JS}[1]{{\color{cyan}{\bf js: #1}}}
\newcommand{\NEW}[1]{{\color{red}{#1}}}

\else
\newcommand{\PF}[1]{{\color{red}{}}}	
\newcommand{\pf}[1]{ #1 }
\newcommand{\HR}[1]{{\color{blue}{}}}
\newcommand{\hr}[1]{#1}%
\newcommand{\VC}[1]{{\color{green}{}}}
\newcommand{\ms}[1]{ #1 }
\newcommand{\MS}[1]{{\color{green}{}}}
\newcommand{\NEW}[1]{#1}
\fi

\newcommand{\va}{\mathbf{a}}
\newcommand{\vb}{\mathbf{b}}
\newcommand{\vd}{\mathbf{d}}
\newcommand{\ve}{\mathbf{e}}
\newcommand{\vf}{\mathbf{f}}
\newcommand{\vg}{\mathbf{g}}
\newcommand{\vh}{\mathbf{h}}
\newcommand{\vi}{\mathbf{i}}
\newcommand{\vj}{\mathbf{j}}
\newcommand{\vk}{\mathbf{k}}
\newcommand{\vl}{\mathbf{l}}
\newcommand{\vm}{\mathbf{m}}
\newcommand{\vn}{\mathbf{n}}
\newcommand{\vo}{\mathbf{o}}
\newcommand{\vp}{\mathbf{p}}
\newcommand{\vq}{\mathbf{q}}
\newcommand{\vr}{\mathbf{r}}
\newcommand{\vt}{\mathbf{t}}
\newcommand{\vu}{\mathbf{u}}
\newcommand{\vv}{\mathbf{v}}
\newcommand{\vw}{\mathbf{w}}
\newcommand{\vx}{\mathbf{x}}
\newcommand{\vy}{\mathbf{y}}
\newcommand{\vz}{\mathbf{z}}

\newcommand{\mA}{\mathbf{A}}
\newcommand{\mB}{\mathbf{B}}
\newcommand{\mC}{\mathbf{C}}
\newcommand{\mD}{\mathbf{D}}
\newcommand{\mE}{\mathbf{E}}
\newcommand{\mF}{\mathbf{F}}
\newcommand{\mG}{\mathbf{G}}
\newcommand{\mH}{\mathbf{H}}
\newcommand{\mI}{\mathbf{I}}
\newcommand{\mJ}{\mathbf{J}}
\newcommand{\mK}{\mathbf{K}}
\newcommand{\mL}{\mathbf{L}}
\newcommand{\mM}{\mathbf{M}}
\newcommand{\mN}{\mathbf{N}}
\newcommand{\mO}{\mathbf{O}}
\newcommand{\mP}{\mathbf{P}}
\newcommand{\mQ}{\mathbf{Q}}
\newcommand{\mR}{\mathbf{R}}
\newcommand{\mS}{\mathbf{S}}
\newcommand{\mT}{\mathbf{T}}
\newcommand{\mU}{\mathbf{U}}
\newcommand{\mV}{\mathbf{V}}
\newcommand{\mW}{\mathbf{W}}
\newcommand{\mX}{\mathbf{X}}
\newcommand{\mY}{\mathbf{Y}}
\newcommand{\mZ}{\mathbf{Z}}

\newcommand{\cA}{\mathcal A}
\newcommand{\cB}{\mathcal B}
\newcommand{\cC}{\mathcal C}
\newcommand{\cD}{\mathcal D}
\newcommand{\cE}{\mathcal E}
\newcommand{\cF}{\mathcal F}
\newcommand{\cG}{\mathcal G}
\newcommand{\cH}{\mathcal H}
\newcommand{\cI}{\mathcal I}
\newcommand{\cJ}{\mathcal J}
\newcommand{\cK}{\mathcal K}
\newcommand{\cL}{\mathcal L}
\newcommand{\cM}{\mathcal M}
\newcommand{\cN}{\mathcal N}
\newcommand{\cO}{\mathcal O}
\newcommand{\cP}{\mathcal P}
\newcommand{\cQ}{\mathcal Q}
\newcommand{\cR}{\mathcal R}
\newcommand{\cS}{\mathcal S}
\newcommand{\cT}{\mathcal T}
\newcommand{\cU}{\mathcal U}
\newcommand{\cV}{\mathcal V}
\newcommand{\cW}{\mathcal W}
\newcommand{\cX}{\mathcal X}
\newcommand{\cY}{\mathcal Y}
\newcommand{\cZ}{\mathcal Z}

\newcommand{\TODO}[1]{\textcolor{cyan}{#1}}

\newcommand{\ST}{\mathcal{T}}
\newcommand{\SST}{\mathcal{T}_S}

\newcommand{\R}{\mathbb{R}}
\newcommand{\Seg}{\mathbf{S}} % geometric part
\newcommand{\Latent}{\mathbf{L}}
\newcommand{\LatentG}{\Latent^{\text{3D}}} % geometric part
\newcommand{\LatentA}{\Latent^\text{app}} % appearance part
\newcommand{\LatentBG}{\mB} % appearance part

\newcommand{\comment}[1]{}

\newcommand{\norm}[1]{\left\lVert#1\right\rVert}
\newcommand{\argmin}{\operatornamewithlimits{argmin}}
\newcommand{\erf}{\operatornamewithlimits{erf}}

\newcommand{\parag}[1]{\vspace{-3mm}\paragraph{#1}}

\newcommand{\ours}[0]{{\bf Ours}}
\newcommand{\direct}[0]{{\bf Resnet}}
\newcommand{\LCR}[0]{{\bf LCR}}
\newcommand{\ECCV}[0]{{\bf NVS-encoder}}
\newcommand{\CVPR}[0]{{\bf Multiview}}
\newcommand{\auto}[0]{{\bf Auto-encoder}}

%% file: tex/abstract.tex
% !TEX root = ../top.tex
% !TEX spellcheck = en-US

\begin{abstract}

Learning general image representations has proven key to the success of many computer vision tasks. For example, many approaches to image understanding problems rely on deep networks that were initially trained on ImageNet, mostly because the learned features are a valuable starting point to learn from limited labeled data. However, when it comes to 3D motion capture of multiple people, these features are only of limited use. 

In this paper, we therefore propose an approach to learning features that are useful for this purpose. To this end, we introduce a self-supervised approach to learning what we call a neural scene decomposition (NSD) that can be exploited for 3D pose estimation. NSD  comprises three layers of abstraction to represent human subjects: spatial layout in terms of bounding-boxes and relative depth; a 2D shape representation in terms of an instance segmentation mask; and subject-specific appearance and 3D pose information. By exploiting self-supervision coming from multiview data, our NSD model can be trained end-to-end without any 2D or 3D supervision. In contrast to previous approaches, it works for multiple persons and full-frame images.
 Because it encodes 3D geometry, NSD can then be effectively leveraged to train a 3D pose estimation network from small amounts of annotated data. 

\end{abstract}

%% file: tex/introduction.tex
% !TEX root = ../top.tex
% !TEX spellcheck = en-US

\section{Introduction}

Most state-of-the-art approaches to 3D pose estimation use a deep network to regress from the image either directly to 3D joint locations or to 2D ones, which are then lifted to 3D using another deep network. In either case, this requires large amounts of training data that may be hard to obtain, especially when attempting to model non-standard motions. 

In other areas of computer vision, such as image classification and object detection, this has been handled by using a large, generic, annotated database to train networks to produce features that generalize well to new tasks. These features can then be fed to other, task-specific deep nets, which can be trained using far less labeled data. AlexNet~\cite{Krizhevsky12} and VGG~\cite{Simonyan15} have proved to be remarkably successful at this, resulting in many striking advances. 

\input{tex/fig_teaser.tex}

Our goal is to enable a similar gain for 3D human pose estimation. A major challenge is that there is no large, generic, annotated database equivalent to those used to train AlexNet and VGG that can be used to learn our new representation. For example, Human 3.6M~\cite{Ionescu14b} only features a limited range of motions and appearances, even though it is one of the largest publicly available human motion databases. Thus, only limited supervision has to suffice.

As a step towards our ultimate goal, we therefore introduce a new scene and body representation that facilitates the training of 3D pose estimators, even when only little annotated data is available. To this end, we train a neural network that infers a compositional scene representation that comprises three levels of abstraction. We will refer to it as \emph{Neural Scene Decomposition} (NSD). As shown in Fig.~\ref{fig:teaser}, the first one captures the spatial layout in terms of bounding boxes and relative depth; the second  is a  pixel-wise instance segmentation of the body; the third is a geometry-aware hidden space that encodes the 3D pose, shape and appearance independently. Compared to existing solutions, NSD enables us to deal with full-frame input and multiple people. As bounding boxes may  overlap, it is crucial to also infer a depth ordering. The key to instantiating this representation is to use multi-view data at training time for self-supervision. This does not require image annotations, only  knowledge of the number of people in the scene and camera calibration, which is much easier to obtain. 

\input{tex/fig_training.tex}

Our contribution is therefore a powerful representation that lets us train a 3D pose estimation network for multiple people using comparatively little training data, as shown in Fig.~\ref{fig:training}. The network can then be deployed in scenes containing several people potentially occluding each other while requiring neither bounding boxes nor even detailed knowledge of their location or scale. This is made possible though the new concept of Bidirectional Novel View Synthesis (Bi-NVS) and is in stark contrast to other approaches based on classical Novel View Synthesis (NVS). These are designed to work with only a single subject in an image crop so that the whole frame is filled~\cite{Rhodin18b} or require two or more views not only at training time but also at inference time~\cite{Eslami18}. 

\NEW{Our neural network code and the new boxing dataset will be made available upon request for research purposes.}

%% file: tex/fig_teaser.tex
% !TEX root = ../top.tex
% !TEX spellcheck = en-US

\begin{figure}
	\centering
	\resizebox{0.9\linewidth}{!}{
	\includegraphics[]{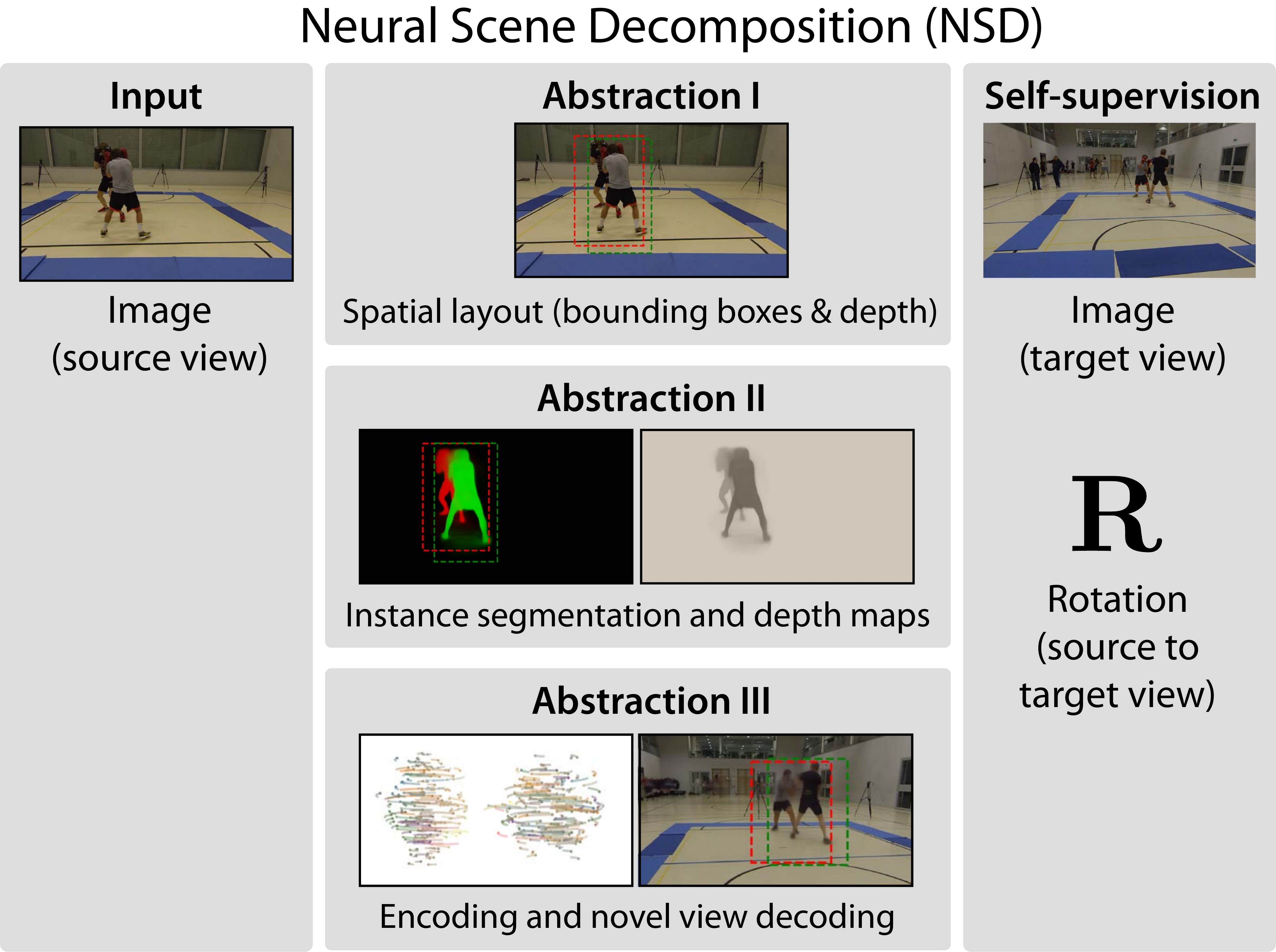}
}
	\caption{\textbf{Neural Scene Decomposition} disentangles an image into foreground and background, subject bounding boxes, depth, instance segmentation, and latent encodings in a fully self-supervised manner using a second view and its relative view transformation for training.}
	\label{fig:teaser}
\end{figure}

%% file: tex/fig_training.tex
% !TEX root = ../top.tex
% !TEX spellcheck = en-US

\begin{figure}
	\centering
	\resizebox{\linewidth}{!}{
	\includegraphics[width=0.5\textwidth]{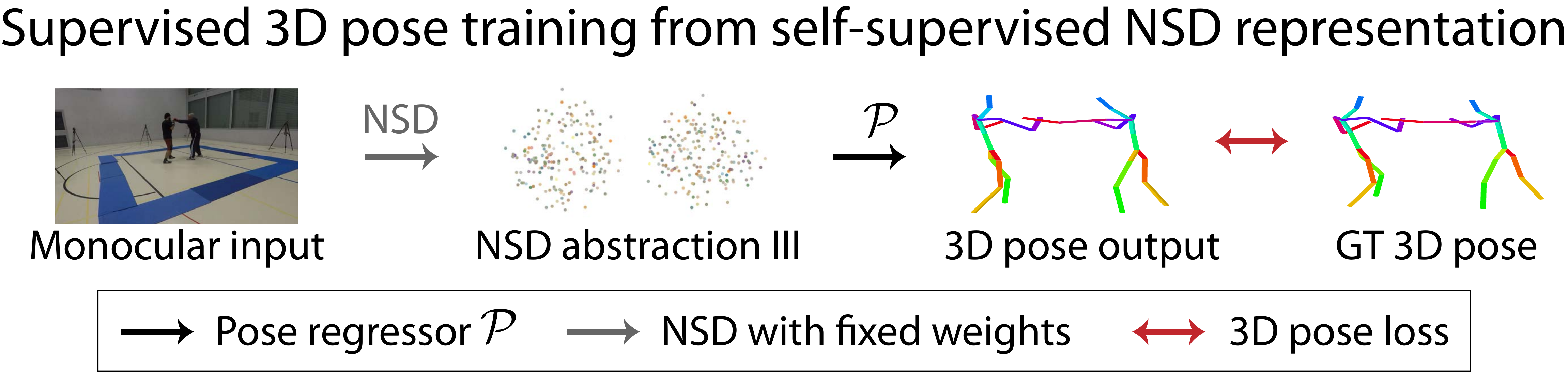}
}
	\caption{\small \textbf{3D pose estimation.} Pose is regressed from the NSD  latent representation that is inferred from the image. During training, the regressor $\cP$ requires far less supervision than if it had to regress directly from the image.}
	\vspace{-5mm}
	\label{fig:training}
\end{figure}

%% file: tex/related.tex
% !TEX root = ../top.tex
% !TEX spellcheck = en-US

\section{Related work}
\label{sec:related}

%Mention POM 1.0 and 2.0

Existing human pose estimation datasets are either large scale but limited to studio conditions, where annotation can be automated \NEW{using marker-less multi–view solutions} \cite{Sigal06,Mehta17a,Ionescu14b,Joo15}, simulated~\cite{Chen16,Varol17}, or generic but small~\cite{Burenius13,Rhodin18a} because manual annotation \cite{Lassner17a} is cumbersome. Multi-person pose datasets are even more difficult to find. Training sets are usually synthesized from single person 3D pose \cite{Mehta18} or multi-person 2D pose \cite{Rogez18} datasets; real ones are tiny and meant for evaluation only \cite{Mehta18,Marcard18}.
In practice, this data bottleneck starkly limits the applicability of deep learning-based single \cite{Pavlakos16,Tome17,Popa17,Moreno17,Martinez17,Mehta17b,Rogez17,Pavlakos17,Zhou17d,Tekin17a,Sun17} and multi-person \cite{Mehta18,Rogez18,Zanfir18a} 3D pose estimation methods. In this section, we review recent approaches to addressing this limitation, in particular those that are most related to ours and exploit unlabeled images for representation learning.

\parag{Weak Pose Supervision.} 
There are many tasks for which labeling is easier than for full 3D pose capture. This  has been exploited via transfer learning~\cite{Mehta17b}, cross-modal variational~\cite{Spurr18} and adversarial~\cite{Yang18b} learning both 2D and 3D pose estimation; minimizing the re-projection error of 3D poses to 2D labels in single~\cite{Zhou17d,Kudo18,Kundu18} and multiple views \cite{Joo15,Pavlakos17};  annotating the joint depth order instead of the absolute position~\cite{PonsMoll14,Pavlakos18a}; re-projection to silhouettes~\cite{Tan17,Tung17self,Kanazawa18b,Pavlakos18,Varol18}.

Closer to us, in~\cite{Simon17}, a 2D pose detector is iteratively refined by imposing view consistency in a massive multi-view studio. A similar approach is pursued in the wild in~\cite{Zhou17a,Rhodin18a}. While effective,  these approaches remain strongly supervised as their performance is closely tied to that of the regressors used for bootstrapping.

In short, all of these methods reduce the required amount of annotations but still need a lot.  Furthermore, the process has to be repeated for new kinds of motion, with potentially different target keypoint locations and appearances. Boxing and skiing are examples of this because they involve motions different enough from standard ones to require full re-training. 
We build upon these methods to further reduce the annotation effort.

\parag{Learning to Reconstruct.}
If multiple views of the same object are available, geometry alone can suffice to infer 3D shape. By building on traditional model-based multi-view reconstruction techniques, networks have been optimized to predict a 3D shape from monocular input that fulfills stereo~\cite{Garg16,Zhou17a}, visual hull~\cite{Yan16,Kar17,Tulsiani18,Rezende16}, and photometric re-projection constraints \cite{Tulsiani17}.
%;usually in form of mesh and voxel representations but also point clouds \cite{Fan17a,Lin17d}. \PF{The sentence after the semi-colon dangles. I am not sure what you mean by it.}
Even single-view training is possible if the observed shape distribution can be captured prior to reconstruction~\cite{Zhu17c,Gadelha16}. 
The focus of these methods is on rigid objects and they do not apply to dynamic and articulated human pose.
%The main drawback of these methods, however, is that they do not apply to the 
% provide semantic information, such as instance segmentation and the desired human 3D joint locations. 
Furthermore, many of them require silhouettes as input, which are difficult to automatically extract from  natural scenes. We address both of these aspects.

\parag{Representation Learning.}
Completely unsupervised methods have been extensively researched for representation learning.  For instance autoencoders have long been used to learn compact image representations~\cite{Bengio12}. %self-supervised: \cite{Doersch15,Wang15c,Agrawal15}.
Well-structured data can also be leveraged to learn disentangled representations, using GANs~\cite{Chen16info,Tran17} or variational autoencoders~\cite{Higgins16}. In general, the image features learned in this manner are rarely relevant to 3D reconstruction.

Such relevance can be  induced by hand-crafting a parameteric rendering function that replaces the decoder in the autoencoder setup~\cite{Tewari17,Bas17}, or by training either the encoder~\cite{Kulkarni15,Grant16,Tung17,Shu17a,Kim17} or the decoder \cite{Dosovitskiy15,Dosovitskiy17} on structured datasets. To encode geometry explicitly, the methods of~\cite{Thewlis17a,Thewlis17b} map to and from spherical mesh representations without supervision and that of~\cite{Zhang18b} selects 2D keypoints to provide a latent encoding. However, these methods have only been applied to well-constrained problems, such as face modeling, and do not provide the hierarchical 3D decomposition we require. 

Most similar to our approach are methods using camera pose estimation \cite{Zamir16} and NVS~\cite{Rhodin18b,Eslami18} as auxiliary tasks for geometry-aware representation learning. In particular, it was shown in~\cite{Eslami18} that reinforcement learning of 3D grasping converges much faster when using NVS features instead of raw images. This, however, was demonstrated only in simulation. In~\cite{Rhodin18b}, NVS is applied in natural scenes for human pose estimation, using a geometry-aware representation based on transforming autoencoders~\cite{Hinton11,Cohen14,Worrall17}. This approach, however, is restricted to images of single humans with tight ground-truth bounding box annotations used at training \emph{and} test time. 
Here, we introduce a hierarchical scene decomposition that allows us to deal with images depicting multiple subjects, without requiring any other information than the multiview images and the camera poses during training, and only single view images at test time.

%% file: tex/method.tex
% !TEX root = ../top.tex
% !TEX spellcheck = en-US

\section{Method}

Our goal is to learn a high-level scene representation that is optimized for 3D human pose estimation tasks, that is, detecting people and recovering their pose from {\it single} images. We refer to this as Neural Scene Decomposition (NSD). To create this NSD, we rely at training time on Novel View Synthesis (NVS) using multiple views and enforcing consistency among the results generated from different views. 

Fig.~\ref{fig:teaser} summarizes our approach. Given a scene containing $N$ people, we want to find $N$ corresponding bounding boxes  $(\vb_i)^N_{i=1}$, segmentation masks  $(\Seg_i)^N_{i=1}$, depth plane estimates  $(z_i)^N_{i=1}$, and latent representation $([\LatentA_i,\LatentG_i])^N_{i=1}$ where  $\LatentA_i$ is a vector representing appearance and $\LatentG_i$ a matrix encoding geometry. Our challenge then becomes training a deep network to instantiate this \NEW{scene} decomposition from images in a completely self-supervised fashion. This means training without bounding boxes, human pose estimates, depth, or instance segmentation labels.

To meet this challenge, we ground NSD on standard deep architectures for \NEW{supervised object detection and representation learning~\cite{Kundu18,Tulsiani18b}} and NVS~\cite{Rhodin18b}, and add new network layers and objective functions to enable self-supervision. In the remainder of this section, we first summarize NVS. We then show how we go from there to NSD, first for a single person and then for multiple. We provide more implementation details in the supplementary material.

\input{tex/fig_NSD_single.tex}

\subsection{Novel View Synthesis}
\label{sec:NVS}

Given two images, $(\mI_v,\mI_{v'})$, of the same scene taken from different viewpoints, NVS seeks to synthesize from $\mI_v$ a novel view $\cF(\mI_v,\mR_{v,v'}, \vt_{v,v'})$ that is as close as possible to $\mI_{v'}$, where $\mR_{v,v'}$ and $\vt_{v,v'}$ are the rotation matrix and translation vector defining the camera motion from $v$ to $v'$. This is typically done by minimizing
\begin{equation}
L(\cF(\mI_v, \mR_{v,v'}, \vt_{v,v'}),\mI_{v'}) \;,
\label{eq:NVSloss}
\end{equation}
where $L$ is an appropriate image-difference metric, such as the $L^2$ norm. This requires static and calibrated cameras, which much less labor intensive to setup than precisely annotating many images with bounding boxes, 3D poses, depth ordering, and instance segmentation. This is one of the main attractions of using NVS for training purposes. 

Previous NVS approaches focused merely on rigid objects \cite{Tatarchenko15,Tatarchenko16,Yang15weakly,Park17,Zhou16c,Flynn16,Cohen14,Worrall17}. 
Methods that synthesize human pose and appearance have used clean silhouettes and portrait images \cite{Zhao17,Zhu18c} and intermediate 2D and 3D pose estimates to localize the person's body parts \cite{Ma17,Lassner17b,Zanfir18b,Si18,Balakrishnan18,Wang18d,Kim18b}. 
% Closely are video-to-video translation methods \cite{Wang18d,Kim18b}. All of these focus on image synthesis quality.
We rely on the approach of~\cite{Rhodin18b} that focuses on representation learning and uses an encoding-decoding architecture without needing human pose supervision. Its encoder $\cE(\cdot)$ maps the input image $\mI_v$ to an appearance vector $\LatentA_v$ and a 3D point cloud $\LatentG_v$ that represents geometry. In the rest of the paper we will refer to the pair $[\LatentA_v,\LatentG_v]$ as the {\it latent representation} of $\mI_v$.  A novel view is then obtained by rotating $\LatentG_v$ by $\mR_{v,v'}$ and then running the decoder $\cD(\cdot)$ on the rotated cloud and original appearance vector, that is, computing $\cD( \mR_{v,v'} \LatentG_v,\LatentA_v)$. 
%\TODO{HR: Put a figure if space permits}

This NVS formulation assumes that subjects are portrayed individually and at the same scale in each image, which makes it possible to ignore the translation $\vt_{v,v'}$ but precludes real-world application where scale may vary significantly. In practice, this is achieved by exploiting the ground-truth bounding box around each subject at both training and test time.
%it requires a bounding box annotation around each subject. 
%\MS{Is the bounding box requirement in addition to the previous assumption or the means to satisfy this assumption?} \HR{Bounding box gives scale and position, not in addition.}

To overcome this limitation, we propose to complement the latent representations produced by this NVS-based scene decomposition with all the information required to deal with multiple people appearing at different scales in the multi-view images. We therefore introduce a novel  architecture that we first describe in the case where there is only one person and then in the multi-person scenario. 

\subsection{NSD with a Single Subject}
\label{sec:NSDSingle}

Existing NVS solutions require scale and position normalization because changes in object scale and translations along the camera optical axis can compensate each other under perspective projection. In particular, a person's absolute height can be predicted from an image only with uncertainty \cite{Gunel18}.  Hence, it is geometrically impossible to predict the size and position in a novel view. 

To alleviate this problem and to attain the sought NSD, we introduce an explicit detection and localization step, along with the notion of bidirectional NVS, that allows us to mix the information extracted from two views in the NVS process. Our complete framework is outlined in Fig.~\ref{fig:NSD_single}.  We now describe each one of these components individually, assuming there is only one person in the scene.

\parag{Subject Detection and Localization.}
To estimate the position and observed size of a single subject whose height and 3D location are initially unknown, we run a detector network $\cB$ on the input image $\mI_v$. Let $\vb_{v}=\cB(\mI_v)$ denote the resulting  bounding box that tightly contains the subject. We use it to define the spatial transformer (ST) network, $\cT$, that returns $\bar{\mI}_{v} = \cT(\mI_v,\vb_v)$,
%\HR{If we feed the bounding box this 'network' has no parameters any more, it is the same for all views. i.e. we can drop the subscript $v$, I did so.} \MS{I am not very knowledgeable about STNs. Is this a correct notation?}, 
an image window of fixed size in which the person is centered. As both detection and windowing are performed by neural networks, this process is end-to-end differentiable.

\parag{Bidirectional NVS.}

The simplest way to use the detections described above would be to obtain them in two views $\mI_{v}$ and $\mI_{v'}$ and apply the NVS strategy of Section~\ref{sec:NVS} to the corresponding windows $\bar{\mI}_{v}$ and $\bar{\mI}_{v'}$, that is, aim to approximate $\bar{\mI}_{v'}$ as $\cF(\bar{\mI}_v, \mR_{v,v'})$. This, however, would provide very little supervisory signal to the detection process and may result in trivial solutions where the detector focuses on background regions that are easy to match. To prevent this, we propose to reconstruct the {\it entire} image $\mI_{v'}$ instead of just the window $\bar{\mI}_{v'}$. This requires mixing the representations of the two views $v$ and $v'$, because generating the entire image $\mI_{v'}$ from the window $\bar{\mI}_{v}$ requires knowing \NEW{the background and }where to insert the transformed version of this window. Therefore, we \NEW{estimate background images} and simultaneously approximate  $\mI_{v'}$ given $\mI_v$ and $\mI_{v}$ given~$\mI_{v'}$.
%Furthermore, to better mix the decompositions obtained from the two views, we reconstruct in a bidirectional manner, that is, by simultaneously approximating  $\mI_{v'}$ given $\mI_v$ and $\mI_{v}$ given $\mI_{v'}$.}
%\HR{The main reason for this mixing is that if the depth in the source is unknown it is impossible to predict the position in the target, using bounding boxes or not. To circumvent this problem we introduce the mixing.}
%\MS{Better?}

%Standard NVS, as described above, synthesizes an image approximating $\mI_{v'}$ given $\mI_{v}$. We expand on this by simultaneously approximating  $\mI_{v'}$ given $\mI_v$ and $\mI_{v}$ given $\mI_{v'}$ by exchanging the subject's location estimates and background images. We assume the cameras to be static and compute background images $\mB_v$ and  $\mB_{v'}$ by taking the median pixel value across all frames in views $v$ and $v'$, respectively.

Formally, given the bounding boxes and spatial transformer introduced above, applying the encoder $\cE$ of Section~\ref{sec:NVS} to both image windows $\bar{\mI}_{v}=\cT(\mI_v,\vb_v)$ and $\bar{\mI}_{v'}=\cT(\mI_v',\vb_{v'})$ returns the latent representations $[\LatentA_{v}, \LatentG_{v}]$ and  $[\LatentA_{v'}, \LatentG_{v'}]$, one per view. We can then invoke the decoder $\cD$ of Section~\ref{sec:NVS} to reconstruct the entire images as
\begin{align}
\hat{\mI}_v  &  = \cT^{-1}(\cD(\LatentA_{v'}, \mR_{v',v} \LatentG_{v'}), \vb_v)  \; , \label{eq:decoding}\\
\hat{\mI}_{v'} & = \cT^{-1}(\cD(\LatentA_{v}, \mR_{v,v'} \LatentG_{v}), \vb_{v'}) \; .  \nonumber
\end{align}
Intuitively, the reconstruction $\hat{\mI}_v$ of view $v$ is obtained by taking the pose seen in $v'$, rotating it to view $v$, 
%\HR{appearance also comes from v', fixed it} 
applying the appearance in view $v'$ to it, and reversing the spatial transformation obtained from view $v$. Equivalently, $\hat{\mI}_{v'}$ is reconstructed from $v$, with the roles of $v$ and $v'$ exchanged. As such, the two reconstructions exchange parts of their decomposition, which creates a bidirectional synthesis. 
%\MS{Please check carefully, because I think the previous version of the equations was wrong. Addition to this comment: After seeing the rest, I now understand that I was confused by the notation. It seems a lot more intuitive to me, however, to have $\hat{\mI}_v$ denote an approximation of $\mI_v$, not of $\mI_{v'}$.} \HR{Before we tried to explain how to get from $I_v$ to $\hat{I}_{v'}$, but this way is also fine. I fixed the appearance origin and ST subscript and brought back a sentence about the opposing direction.}
%$\hat{\mI}_v$ depicts the subject with the appearance and pose seen in $v$ but rotated and translated to the camera direction and image location estimated in $v'$. 
%We reverse the roles of $v$ and $v'$ to compute for $\hat{\mI}_{v'}$ and $\bar{\mS}_{v'}$, which creates a bidirectional synthesis with shared location information.

%One problem arising from working with image windows extracted from bounding boxes is that they contain some background information. Ignoring this, as above, translates to having complete image reconstructions that are contaminated in places by the wrong background. 
\NEW{The final ingredient is to blend in the target view background.}
To \NEW{make this easier}, we assume the cameras to be static and compute background images $\mB_v$ and  $\mB_{v'}$ by taking the median pixel value across all frames in views $v$ and $v'$, respectively. 
%\NEW{By this assumption, we can train the detector to focus on the foreground---everything not contained in background image and visible from multiple views.}
For each view, we then learn to produce a segmentation mask $\bar{\mS}_{v}$ 
%\MS{Since we don't have any ground-truth masks, why do we put a hat on S? Can't we just use $\mS_{v}$?}\HR{Well in principle one could compare to a GT segmentation for evaluation, but I have not planned that yet. Was just to be consistent with I.} 
as an additional output channel of the decoder $\cD$. Since this mask corresponds to the image window $\bar{\mI}_{v}$, we apply the inverse spatial transformer to obtain a mask $\hat{\mS}_{v'}$ corresponding to the full image. We then use these segmentation masks to blend the reconstructed images $\hat{\mI}_v$ and $\hat{\mI}_{v'}$ of Eq.~\ref{eq:decoding} with the corresponding backgrounds $\mB_v$ and $\mB_{v'}$ to produce the final reconstructions
\begin{align}
\cF_{\mI_{v}}(\mI_{v'}, \mR_{v',v}) &= \hat{\mS}_{v} \hat{\mI}_{v} + (1-\hat{\mS}_{v}) \mB_{v} \nonumber\\
\cF_{\mI_{v'}}(\mI_{v}, \mR_{v,v'}) &= \hat{\mS}_{v'} \hat{\mI}_{v'} + (1-\hat{\mS}_{v'}) \mB_{v'} \; .
\label{eq:blending}
\end{align}
%\MS{In fact, these equations are wrong. A segmentation mask is defined over the windows $\bar{\mI}_v$, not the entire image. There should therefore be some form of inverse spatial transform applied to them.}
%\HR{I considered it before as an additional channel of $\hat{I}$, which goes indeed together through the spatial transformer.}
%
While our approach to blending is similar in spirit to that of~\cite{Balakrishnan18}, it does not require supervised 2D pose estimation.
It also differs from that of~\cite{Rhodin18b} where the background composition is formulated as a sum without explicit masks. The generated segmentation masks allows NSD to operate on images with complex background at test time and equips it with a shape abstraction layer. 

\input{tex/fig_NSD_double.tex}

\subsection{NSD with Multiple Subjects}
\label{sec:NSDMultiple}

The approach of Section~\ref{sec:NSDSingle} assumes that there is a single subject in the field of view. We now extend it to the case where there are a fixed number of $N>1$ subjects of varying stature. To this end, we first generalize the detector $\cB$ to produce $N$ bounding boxes $(\vb_{v,i})_{i=1}^{N}=\cB(I_v)$, instead of only one. NVS is then applied in parallel for each detection as shown in Fig.~\ref{fig:NSD_double}. This yields tuples of latent codes $(\LatentA_{v,i})^{N}_{i=1}$ and $(\LatentG_{v,i})^{N}_{i=1}$, 
%synthesized images 
transformed windows $(\bar{\mI}_{v',i})^{N}_{i=1}$, and corresponding segmentation masks $(\bar{\mS}_{v',i})^{N}_{i=1}$. The only step that is truly modified with respect to the single subject case is the compositing of Eq.~\ref{eq:blending} that must now account for potential occlusions. This requires the following two extensions. 
%\PF{What happens when there are fewer than $N$ people in the scene? What are typical values for $N$?} \HR{I tested with N=3 when there is just 1 or 2 persons. It does not work all too well. I have not tested it for more than 2 persons, would need a different network architecture to be memory efficient and more GPUs to train such complex models on.}

\parag{Appearance-based view association.}

\input{tex/fig_association.tex}

Objects in the source and target views are detected independently. To implement the bidirectional NVS of Section~\ref{sec:NVS}, we need to establish correspondences between bounding boxes in views $v$ and $v'$. Doing so solely on the basis of geometry would result in mismatches due to depth ambiguities. To prevent this, we perform an appearance-based matching.  As shown in Fig.~\ref{fig:association}, it relies on the fact that the appearance latent vector $\LatentA_{v,i}$ of object $i$ in view $v$ should be the same same as $\LatentA_{v',j}$ in view $v'$ when $i$ and $j$ correspond to the same person in views $v$ and $v'$.
%, and different otherwise. 
We therefore build the similarity matrix $\mM$ whose elements are the cosine distances
\begin{equation}
\mM_{j,i} = \frac{\LatentA_{v,i} \cdot \LatentA_{v',j} }{||\LatentA_{v,i}||\  ||\LatentA_{v',j}||} ,
\end{equation}
where $\cdot$ is the dot product. In practice, we \NEW{found that using} only the first 16 out of 128 latent variables of the $\LatentA_{v,i}$s in this operation to leave room to encode commonalities between different subjects in $\LatentA$ \NEW{for the NVS task} while still allowing \NEW{for distinctive similarity matrices for the purpose of association}. Ideally, subject $i$ in view $v$ should be matched to the subject $j^*$ in view $v'$ for which $\mM_{j,i}$ is maximized with respect to $j$. To make this operation differentiable, we apply a row-wise softmax of the scaled similarity matrix $\beta \mM$, with $\beta=10$  to promote sharp distinctions. We use the resulting $N\times N$ association matrix $\mA$ to re-order the 
%synthesized object views 
transformed windows
and segmentation masks as
\begin{align}
(\bar{\mI}_{v',j})^{N}_{j=1}  & \leftarrow \mA (\bar{\mI}_{v',i})^{N}_{i=1}\; ,\nonumber\\
(\bar{\mS}_{v',j})^{N}_{j=1} & \leftarrow \mA (\bar{\mS}_{v',i})^{N}_{i=1} \; .
\label{eq:association}
\end{align}
This weighted permutation is differentiable and, hence, enables end-to-end training. 
%\PF{Weighted permutation? So this is a differentiable Hungarian algorithm? What happens for empty detections?} \HR{Such outliers are still matched to the closes one. At training time the network is supposed to change the detector to avoid this.. One could think about a gating unit in case no similar person is found. Or possibly solving some MRF like Pierre did.}
%\MS{In essence, we assume that we always observe a fixed and given number of people $N$, correct? If so, should we state this explicitly at the beginning of this section, so as to avoid the reviewers being confused similarly to Pascal?}
%\HR{Yes, I think this is most realistic. Im running testes where N> the true number of persons, but it does not work very well..}

\parag{Reasoning about Depth.}

\input{tex/fig_depth.tex}
After re-ordering the transformed windows and segmentation masks, the reconstructed image for view $v$ can in principle be obtained as
\begin{equation}
\cF_{\mI_{v}}((\mI_{v',i}), \mR_{v',v}) = \left(\sum_{i=1}^N\hat{\mS}_{v,i} \hat{\mI}_{v,i}\right) + \left(1-\sum_{i=1}^N\hat{\mS}_{v,i}\right) \mB_{v}\;,
\end{equation}
where $\hat{\mI}_{v,i}$ is the initial reconstruction for view $v$ and subject $i$, computed independently for each person via Eq.~\ref{eq:decoding}. In short, this combines the foreground region of every subject with the overall background. This strategy, however, does not account for overlapping subjects; depth is ignored when computing the intensity of a pixel that is covered by two foreground masks.

%When objects overlap, proper depth ordering should be ensured and it should be done in a differentiable way. With this goal in mind, 
To address this, we extend the detector $\cB$ to predict a depth value $z_{v,i}$ in addition to the bounding boxes. We then compute a visibility map for each subject based on the depth values of all subjects and their segmentation masks. To this end, we use the occlusion model introduced in~\cite{Rhodin15,Rhodin16b} that approximates solid surfaces with Gaussian densities to attain differentiability. This model relies on the transmittance to depth $z$, given in our case by $\mT(z) = \exp( - \sum_i \mS_{v,i} (\erf(z_{v,i}-z)+1))$. Given this transmittance, the visibility of subject $i$ is then defined as $\mT(z_{v,i})\mS_{v,i}$. \NEW{These visibility maps form the instance segmentation masks, and we obtain depth maps by weighting each by $z_{v,i}$.} This process is depicted in Fig.~\ref{fig:depth}. Altogether, this lets us re-write the reconstruction of image $\mI_v$ as
\begin{equation}
\cF_{\mI_{v}}((\mI_{v',i}), \mR_{v',v}) = \left(\sum_i \mT(z_{v,i})\mS_{v,i} \hat{\mI}_{v,i}\right) Z + \mT(\infty) \mB_v\;,
\label{eq:visibility}
\end{equation}
where $Z = \frac{1-\mT(\infty)}{\sum_j\mT(z_{v,j})\mS_{v',i}}$ is a normalization term. More details on this occlusion model are provided in the supplementary material.

%\paragraph{Discussion.}
If at all, depth order in NVS has been handled through depth maps \cite{Tatarchenko15} and by introducing a discrete number of equally spaced depth layers \cite{Flynn16}, but none of these address the inherent scale ambiguity as done here with Bi-NVS. Closely related is the unsupervised person detection and segmentation method proposed in \cite{Baque17b}, which localizes and matches persons across views through a grid of candidate positions on the ground plane.
%, which however requires the presence and knowledge of that plane. \MS{This could go.}

\NEW{In short, we train a combined detetection-encoding-decoding network to individually detect, order, and model foreground objects, that is, the objects visible from all views and not contained in the static background.}

\subsection{NSD Training}
\label{sec:nsdTraining}

NSD is trained in a fully self-supervised fashion to carry out Bi-NVS as described in Section~\ref{sec:NSDSingle}. We perform gradient descent on batches containing pairs of images taken from two or more available views at random. Since no labels for intermediate supervision are available and $\cB$, $\cE$ and $\cD$ are deep neural networks, we found end-to-end training to be unreliable and rely on the following. \NEW{To counteract, we introduce focal spatial transformers (explained in the supplemental document) and the following priors.}

 \parag{Using Weak Priors.}
 \label{sec:prior}
 
 Without guidance, the detector converged to a fixed location on an easy to memorize background patch. To push the optimization process towards exploring detection positions on the whole image, we add a loss term that penalizes the squared deviation of the average bounding box position across a batch from the image center. Note that this is different from penalizing the position of each detection independently, which would lead to a strong bias towards the center. 
 Instead, it assumes a Gaussian prior on the average person position, which is fulfilled not only when the subjects are normally distributed around the center, but, by the central limit theorem, also when they are  uniformly distributed. We build independent averages for the N detection windows, which avoids trivial solutions.

Similarly, we introduce a scale prior that encourages the average detection size to be close to 0.4 times the total image size and favors an aspect ratio of 1.5. As for position, this prior is weak and would be fulfilled if sizes vary uniformly from 0.1 to 0.7. Both terms are given a small weight of $0.1$ to reduce the introduced bias.

%% file: tex/fig_NSD_single.tex
% !TEX root = ../top.tex
% !TEX spellcheck = en-US

\begin{figure*}
	\centering
	\resizebox{\linewidth}{!}{
	\includegraphics[width=0.5\textwidth]{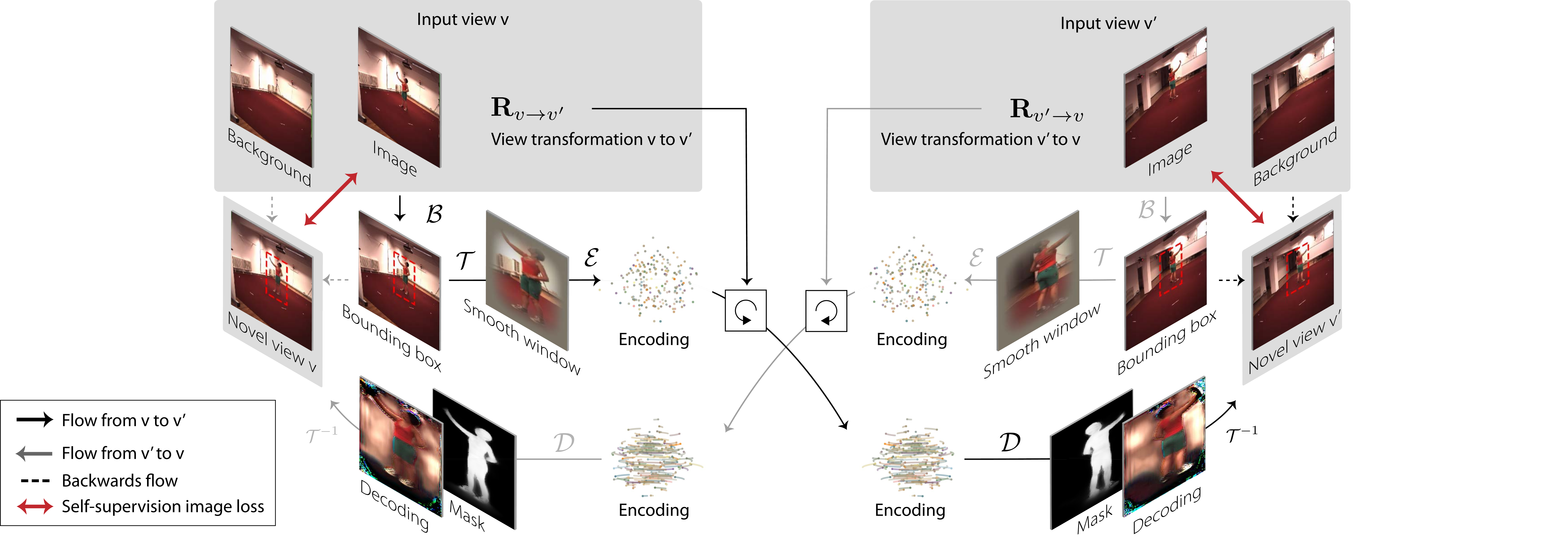}
}
	\caption{\textbf{Bidirectional NVS} jointly predicts a novel view $v'$ from $v$ and $v$ from $v'$, mixing object location and scale estimates between the two directions. This overcomes scale ambiguities in classical NVS, which predicts $v'$ from $v$ without backwards flow from $v'$.}
	\label{fig:NSD_single}
\end{figure*}

%% file: tex/fig_NSD_double.tex
% !TEX root = ../top.tex
% !TEX spellcheck = en-US

\begin{figure*}
	\centering
	\resizebox{\linewidth}{!}{
	\includegraphics[width=0.5\textwidth]{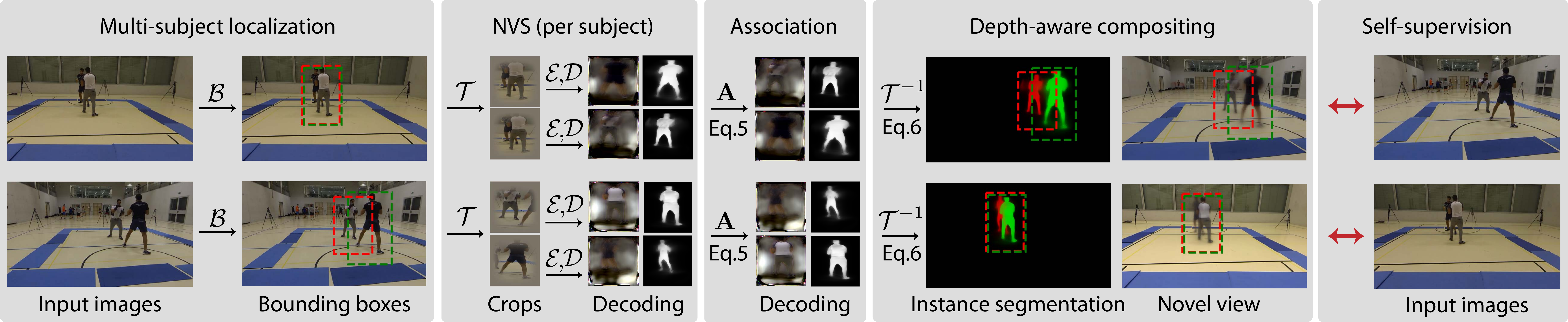}
}
	\caption{\textbf{Multi-person Bi-NVS}. Multiple subjects are detected in each input image and their encoding and decoding is processed separately, akin to single person Bi-NVS. Key is the association of multiple persons across views by Eq.~\ref{eq:association} and their composition by Eq.~\ref{eq:visibility}.
	%	on the background that maintains proper depth ordering. It is explained in detail in Fig.~\ref{fig:depth}.
	}
	\label{fig:NSD_double}
\end{figure*}

%% file: tex/fig_association.tex
% !TEX root = ../top.tex
% !TEX spellcheck = en-US

\begin{figure}[t!]
	\centering
	%\resizebox{0.5\linewidth}{!}{
	\includegraphics[width=1\linewidth]{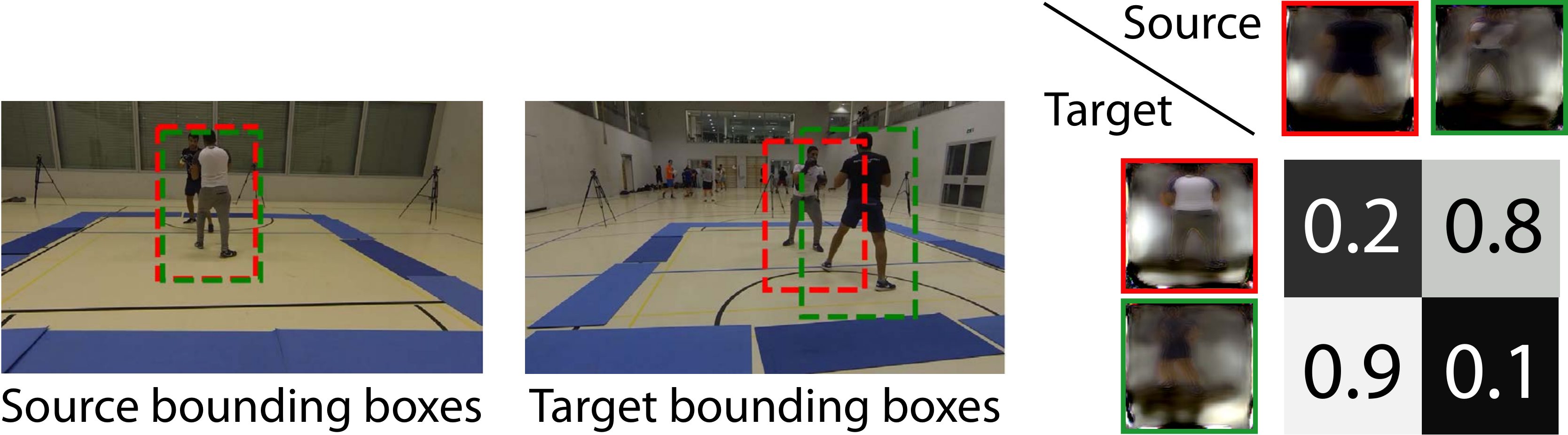}
%}
	\caption{\textbf{Identity association.} In this example, the light subject is detected once as the first and once as the second subject, here visualized by red and green boxes. To match subjects across views, we build a similarity matrix from their respective appearance encodings, as shown on the right. }
	\label{fig:association}
\end{figure}

%% file: tex/fig_depth.tex
% !TEX root = ../top.tex
% !TEX spellcheck = en-US

\begin{figure}
	\centering
	\resizebox{\linewidth}{!}{
	\includegraphics[width=0.5\textwidth]{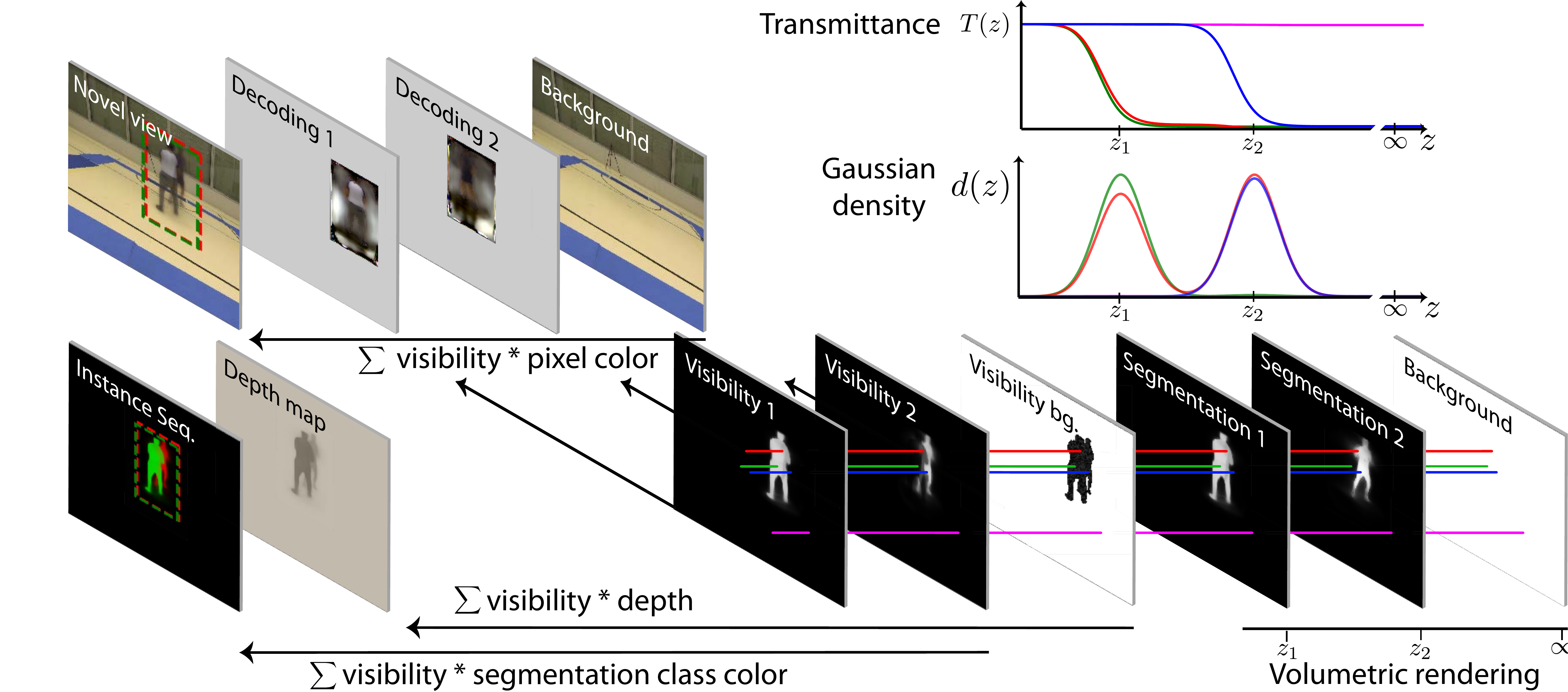}
}
	\caption{\textbf{A visual breakup of Eq.~\ref{eq:visibility}.} The novel view is the sum of decoding layers and background, weighted by their respective visibility maps. Similarly, the segmentation mask and depth map is computed from weighted color and depth values. 
	Visibility is computed through volumetric rendering. We plotted the computation for four pixels marked in red, green blue and magenta. Each person forms a translucent layer with Gaussian density in depth direction (see lower plot), so that transmittance decays smoothly at each layer and proportionally to the segmentation mask (top plot).}
	\label{fig:depth}
\end{figure}

%% file: tex/evaluation.tex
% !TEX root = ../top.tex
% !TEX spellcheck = en-US

\section{Evaluation}
\label{sec:eval}

In this section, we evaluate NSD for the tasks of multi-people detection, 3D pose estimation, and novel view synthesis. 
First, we show that, in single-person scenarios, our method delivers similar accuracy compared to existing self-supervised approaches, even though they require ground-truth bounding box annotations whereas we do {\it not.} Second, we use a boxing scenario that stumps state-of-the-art algorithms to demonstrate that our method can effectively handle closely interacting people. Finally, we provide results on scenes containing three people. Additional scene decomposition and re-composition results are given in the supplementary material.

\subsection{Baselines} 
We refer to our method as \ours{} and compare it against:
\begin{itemize}

 \vspace{-1mm}
 \item \LCR-H36M and \LCR{}-ITW. They are both versions of \LCR{}++~\cite{Rogez18}, which is the current state of the art in multi-person 3D pose estimation. The first is trained on Human3.6M (H36M) and the second on in-the-wild 2D and 3D datasets.
  
% \vspace{-1mm}
% \item \CVPR{}~\cite{Rhodin18a}. A method that exploits consistency across multiple views to provide a supervisory signal. Since it requires closely cropped bounding boxes around the subjects, the published results that we reproduce here are on ground-truth bounding boxes. 
 
%\vspace{-1mm}
%\item \ECCV{}~\cite{Rhodin18b}. A method that, like us, uses NVS for self-supervision but also relies on ground-truth bounding boxes for its published results. 
 
\vspace{-1mm}
\item \direct-$\mI$ and \direct-$\bar{\mI}$. Two baselines that use a \direct{}~\cite{Mehta17b}, whose architecture is similar to the one we use, to regress directly from the image to the 3D pose. \direct-$\mI$ runs on the whole image $\mI$ whereas \direct-$\bar{\mI}$ runs on the cropped one $\bar{\mI}$ that NSD returns. 
 
\vspace{-1mm}
 \item \auto{}. A baseline that uses the same spatial transformer and encoder-decoder as we do but learns an image auto-encoding instead of NVS in \ours{}.

\end{itemize}
In the above list, we distinguish between baselines  \direct-$\mI$, \direct-$\bar{\mI}$, and \auto{} that we implemented ourselves and the recently published method \LCR{}. The latter has been discussed in Section~\ref{sec:related}. We have used publicly available code to run \LCR{} on our data.
% and give the published performance numbers, when available, for \CVPR{} and \ECCV{}. 
 
%\parag{Dataset and Metrics.} 
%
%We test single-person NSD on the PoseTrack2018 challenge of the well known H36M~\cite{Ionescu14b} dataset. The images were recorded in a four-camera studio and the task is to estimate 17 3D joint locations relative to the subject's hip. Accuracy is usually measured in terms of the mean per joint position error (MPJPE). To compare against~\cite{Rhodin18a}, we also report the N-MPJPE, that is, the MPJPE after rigidly aligning the prediction to the ground truth in the least squares sense.

%To test the performance of NSD in the presence of closely interacting people, we introduce a new boxing dataset that comprises 8 sequences with sparing fights between 11 different boxers. We used Captury's markerless motion capture technology\footnote{http://thecaptury.com} to annotate 6 of these sequences. Four form the supervised training set and two are reserved for validation and testing. We use the remaining two unlabeled sequences for semi-supervised representation learning. For evaluation purposes, the predicted and ground-truth poses are matched to minimize error.
% their left-to-right order in the image
%For this dataset, we report the MPJPE, NMPJPE and in addition the rate of detection, which tallies complete misses and wrong matches with an MPJPE over 30cm.
%H36m subsampling: [1, 2, 7, 71, 358]

\input{tex/fig_eval_h36m_qualitative.tex}

\input{tex/fig_eval_h36m.tex}

\subsection{Supervised Training}
\label{sec:supervised}

Recall from Section~\ref{sec:nsdTraining} that we learn our NSD representation in a completely self-supervised way, as shown on the left side of Fig.~\ref{fig:training}. This being done, we can feed an image to the encoder $\cE$ that yields a representation in terms of one or more bounding boxes, along with the corresponding segmentation masks, and latent representations. As our central goal is to demonstrate the usefulness of this representation for 3D pose estimation using comparably little annotated data, we then use varying amounts of such data to train a new network $\cP$ that regresses from the representation to the 3D pose. The right side of Fig.~\ref{fig:training} depicts this process. 

At inference time on an image $\mI$, we therefore compute $\cE(\mI)$ and run the decoder $\cP$ on each resulting bounding box and corresponding latent representation. Because the learned representation is rich, we can use a simple two-layer fully-connected network for $\cP$.

\subsection{Single-Person Reconstruction}

We test single-person NSD on the PoseTrack2018 challenge of the well known H36M~\cite{Ionescu14b} dataset. The images were recorded in a four-camera studio and the task is to estimate 17 3D joint locations relative to the subject's hip. Accuracy is usually measured in terms of the mean per joint position error (MPJPE) expressed in mm. To compare against \cite{Rhodin18a} and \cite{Rhodin18b}, we also report the N-MPJPE, that is, the MPJPE after rigidly aligning the prediction to the ground truth in the least squares sense.

We learn our NSD representation from the training sequences featuring five different subjects. We evaluate on the validation sequences that feature two different subjects.
In Fig.~\ref{fig:h36m_NVS}, we use one image pair from the validation set to show that NSD localizes and scale normalizes a subject well enough for resynthesis in a different view. We provide additional examples in the supplementary material. 

$\cP$ is learned on subsets of the complete training set. In Fig.~\ref{fig:h36m_quantitative}, we plot the MPJPE as a function of the amount of labeled training data we used for supervised training, as described in Section~\ref{sec:supervised}. In practice, the smaller training sets are obtained by regularly sub-sampling the dedicated 35k examples. Direct regression from the full-frame image as in \direct{} is very inaccurate. Using the NSD  bounding boxes as in \direct-$\bar{\mI}$ and \auto{} significantly improves performance. Using our complete model further improves accuracy by exploiting the learned high level abstraction. It remains accurate when using as few as 1\% of the available labels. Fig.~\ref{fig:h36m_pose} depicts predictions obtained when $\cP$ has been trained using less than 15\% of the available labels.

\input{tex/fig_eval_h36m_qualitative_pose.tex}

Among the semi-supervised methods, \ours{} is more than 15mm more accurate than \auto{}. The results reported by \cite{Rhodin18a} and \cite{Rhodin18b} are not directly comparable, since their evaluation is on a non-standard training and test sets of H36M and they use ground truth bounding boxes. Nevertheless, their reported N-MPJPE are higher than ours throughout, for example $153.3$ and $117.6$ for 15k supervised labels while we obtain $91$. This confirms that our approach can handle full-frame input without loosing accuracy.

To demonstrate that our approach benefits from additional multi-view data {\it without} additional annotations, we retrained the encoder $\cE$ using not only the training data but also the  PoseTrack challenge test data for which the ground-truth poses are not available to us. Furthermore, our approach can also be used in a transductive manner, by additionally incorporating the images used during evaluation without the corresponding annotations at training time. We refer to these two strategies as \ours-extended and \ours-transductive, respectively. As can be seen in Fig.~\ref{fig:h36m_NVS}, they both increase accuracy. More specifically, when using only 500 pose labels, the error reduces by 5mm with the former and another 10mm with the latter, as shown in Fig.~\ref{fig:h36m_size}, %\PF{It can also be seen in Fig.~\ref{fig:h36m_NVS}, which makes Fig.~\ref{fig:h36m_size} kind of redundant.}

Of course, many existing methods attain a higher accuracy than \ours{} by using all the annotated data and adding to it either synthetic data or additional 2D pose datasets for stronger supervision. While legitimate under the PoseTrack challenge rules, it goes against our aim to reduce the required amount of labeling. For example, \LCR-H36M reports an accuracy of 49.4mm, but this has required creating an additional training dataset of 557,000 synthetic images to supplement the real ones. Without it, the original \LCR~\cite{Rogez17} achieves accuracies that are very close to those of \ours{}---ranging from 75.8 to 127.1 depending on the motion---when using full supervision. However, the strength of \ours{} is that its accuracy only decreases very slowly when reducing the amount of annotated data being used. 

\input{tex/fig_eval_h36m_unsupervised.tex}

\subsection{Two-Person Reconstruction}

\input{tex/fig_eval_box_qualitative_pose.tex}
\input{tex/fig_eval_boxing.tex}

To test the performance of NSD when two people are interacting, we introduce a new boxing dataset that comprises 8 sequences with sparring fights between 11 different boxers. We used a semi-automated motion capture system~\cite{Captury} to annotate 6 of these sequences, of which we set aside 4 for supervised training of $\cP$ and 2 for testing purposes. We then use the remaining 2  in combination with the annotated training sequences for self-supervised NSD learning. 

Fig.~\ref{fig:box_pose} depicts 3 different images and the recovered 3D poses for each boxer, which are accurate in spite of the strong occlusions. In Fig.~\ref{fig:boxing_quantitative}, we compare our results to those of \LCR-H36M{}, \LCR{}-ITW{}, and \direct-$\bar{\mI}$. We clearly outperform all three. \NEW{While LCR is trained on another dataset, which precludes a direct comparison, this} demonstrates the importance of domain specific training and NSD's ability to learn a depth ordering, which is essential to properly handle occlusions. 

\subsection{Multi-Person Reconstruction}

\input{tex/fig_threePersons.tex}

Our formalism is designed to handle a pre-defined yet arbitrary number of people. To test this, we captured a 10 minute five-camera sequence featuring 6 people interacting in groups of three and used it to train a 3-people NSD representation, still in a fully self-supervised way. Fig.~\ref{fig:threePeople} depict the NSD representation of two images of that sequence, along with the image re-synthesized using it. Note that, in both cases, there are only three people in the re-synthesized image, which makes sense in this case. 

%% file: tex/fig_eval_h36m_qualitative.tex
% !TEX root = ../top.tex
% !TEX spellcheck = en-US

\begin{figure}
	\centering
	%\resizebox{\linewidth}{!}
	\includegraphics[width=0.6\linewidth]{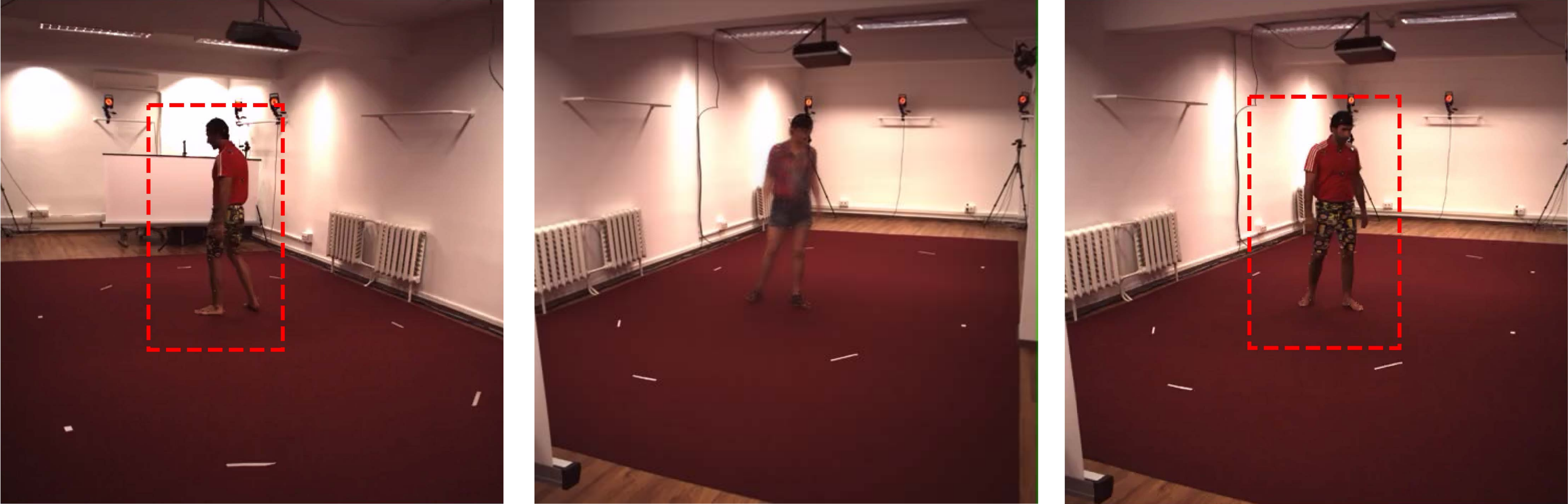}
	\caption{\textbf{Novel view synthesis.} The images on the left and right were taken at the same time by two different cameras. The dotted lines denote the NSD bounding box. The image in the middle was synthesized from the image on the left with the subject retaining his original appearance, with shorts instead of long pants, but being shown in the pose of the one on the right. %\PF{Are they the same person? }
	}
	\label{fig:h36m_NVS}
\end{figure}

%% file: tex/fig_eval_h36m.tex
% !TEX root = ../top.tex
% !TEX spellcheck = en-US

\begin{figure}
	\centering
	\resizebox{\linewidth}{!}{
	\includegraphics[width=0.5\textwidth]{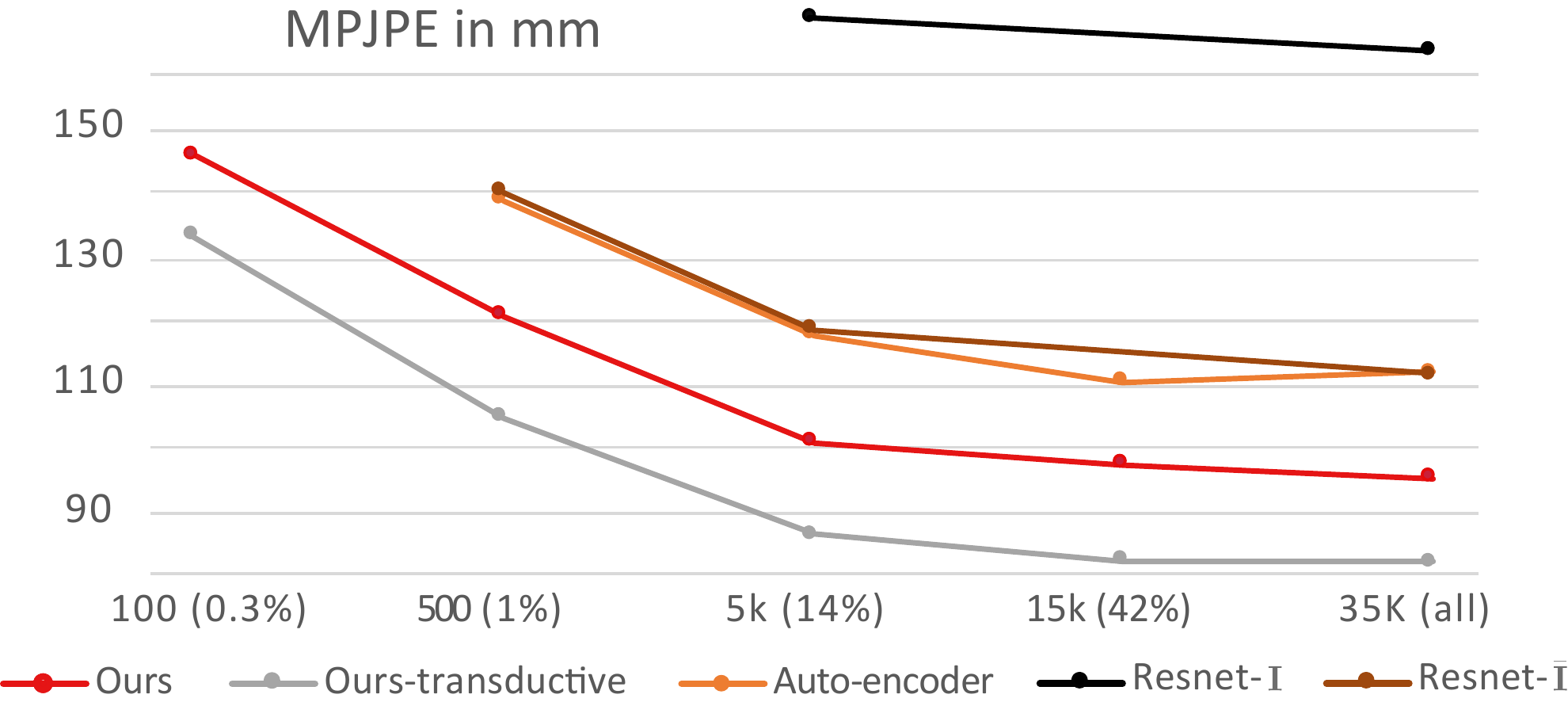}
}
	\caption{\small \textbf{Accuracy of single-person NSD.} We plot the MPJPE  on the PoseTrack validation set as a function of the number of training samples  used to train $\cP$.}
	\label{fig:h36m_quantitative}
\end{figure}

%Our method is the most accurate, compared to direct regression from images, image representation learning, and existing self-supervised methods, despite handling the more difficult full-frame input without bounding box annotation. $^\star$The related work marked with dashed lines used a slightly different train/val/test split and N-MPJPE, which might cause minor differences.

%% file: tex/fig_eval_h36m_qualitative_pose.tex
% !TEX root = ../top.tex
% !TEX spellcheck = en-US

\begin{figure}
	\centering
	\includegraphics[width=1\linewidth]{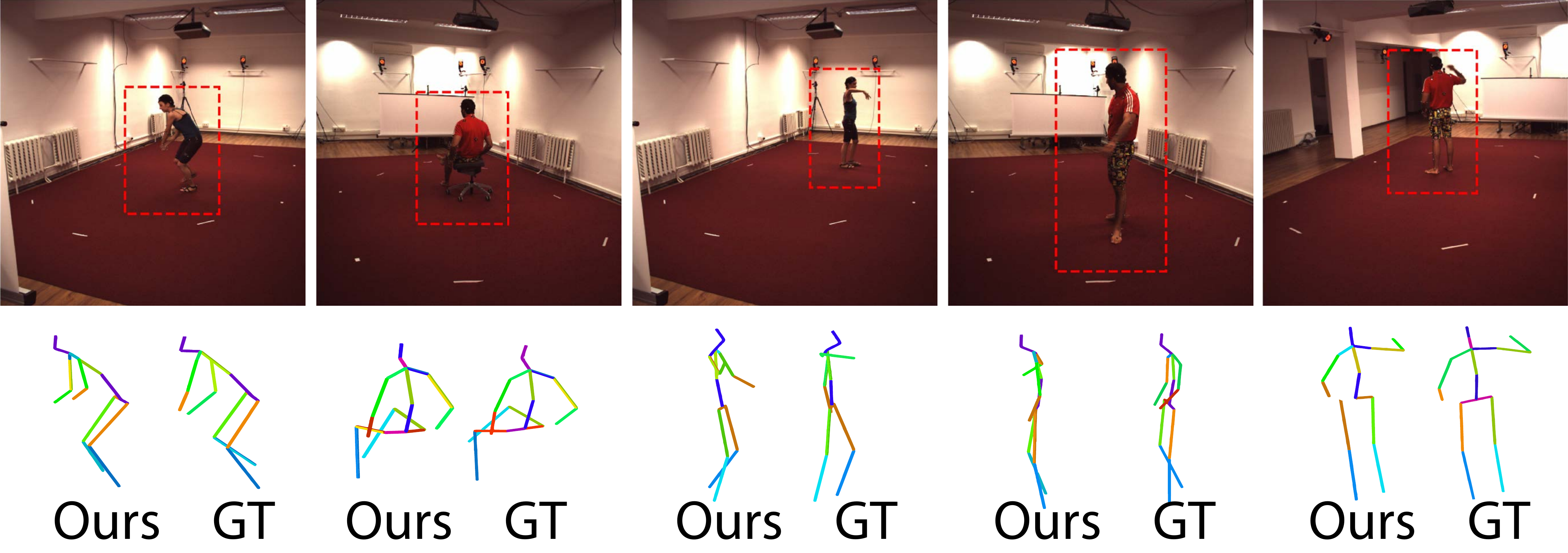}
	\caption{\small \textbf{Pose estimation} using only 15\% of the training labels to train $\cP$. Top row: Images with detected bounding box. Bottom row: Recovered and ground-truth poses shown side by side.}
	\label{fig:h36m_pose}
\end{figure}

%% file: tex/fig_eval_h36m_unsupervised.tex
% !TEX root = ../top.tex
% !TEX spellcheck = en-US

\begin{figure}
	\centering
	%\resizebox{\linewidth}{!}
	{
	\includegraphics[width=0.5\linewidth]{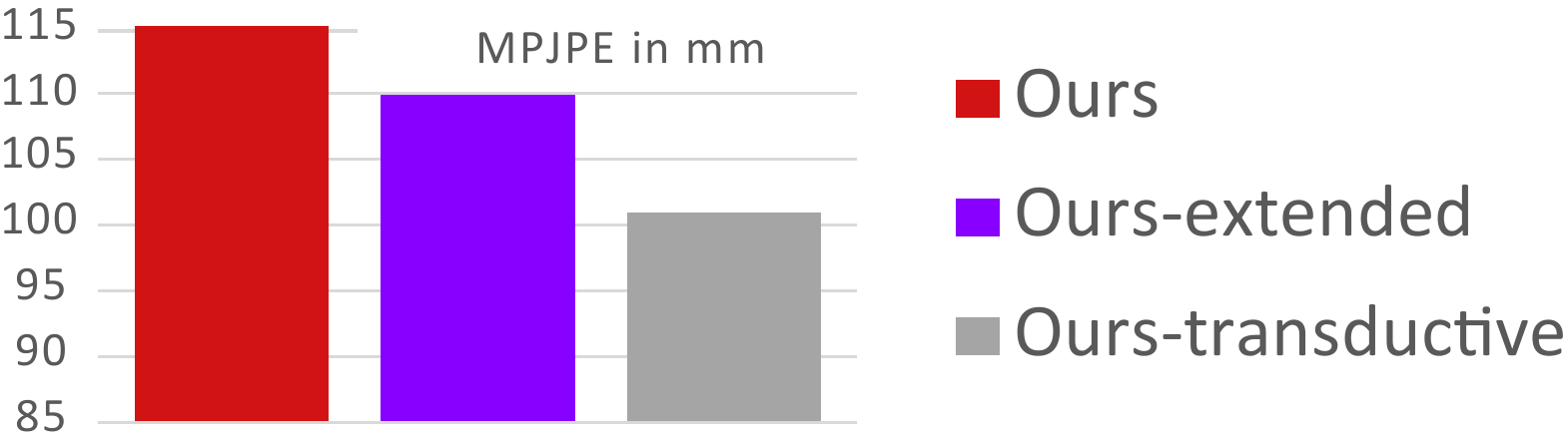}
}
	\caption{\small \textbf{Varying the unlabeled training set size.} Using more samples improves accuracy, particularly in the transductive case, where examples come from the test distribution.}
	\label{fig:h36m_size}
\end{figure}

%% file: tex/fig_eval_box_qualitative_pose.tex
% !TEX root = ../top.tex
% !TEX spellcheck = en-US

\begin{figure}
	\centering
	%\resizebox{\linewidth}{!}
	\includegraphics[width=1\linewidth]{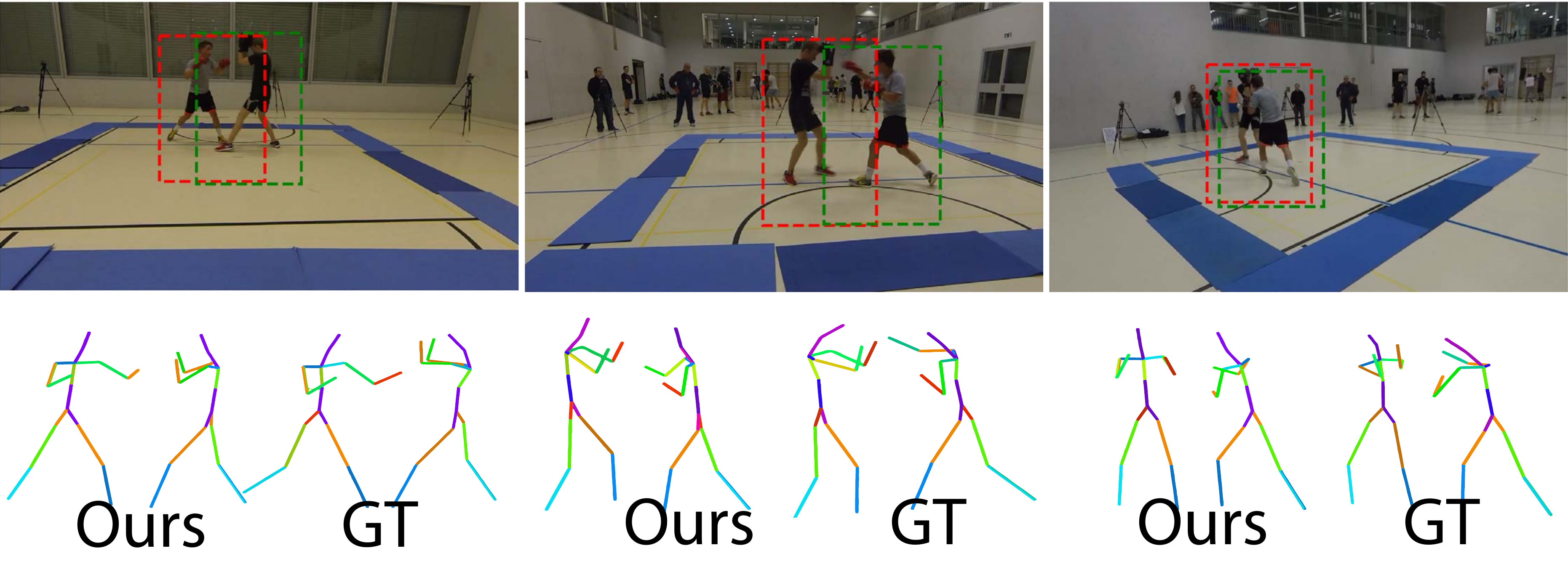}
	\caption{\small \textbf{Estimating the poses of two boxers.} Top row: Images with one detected bounding box per boxer. Bottom row: Recovered and ground-truth poses shown side by side.}
	\label{fig:box_pose}
\end{figure}

%% file: tex/fig_eval_boxing.tex
% !TEX root = ../top.tex
% !TEX spellcheck = en-US

\begin{figure}
	\centering
\resizebox{1\columnwidth}{!}%
{%
	\begin{tabular}[b]{ |l|c|c|c|c| }%
		%			\multicolumn{1}{r}{}
		%			&  \multicolumn{3}{c}{MPJPE reconstruction error [mm]}\\
		%\multicolumn{5}{c}{Supervised training on all subjects of H36M}\\
		\hline
		Method & MPJPE in mm & NMPJPE in mm & NMPJPE$^\star$ in mm& Detection rate \\
		\hline
		\ours{} & \textbf{125.4} & \textbf{99.7} & \textbf{97.8} & \textbf{99.8} \% \\
		\LCR{}-ITW{} & 155.6 & 154.37 & 122.7 & 79.7 \% \\ % detection rate on MPJPE: 100-20.4
%		LCR-ITW$^\star$ & 125.8 & 122.7 & 79.7 \% \\ % detection rate on MPJPE: 100-20.4
%		LCR-ITW & 155.6 & 154.37 & 100-21.8\% \\ % without normalization
%		LCR-Demo$^\star$ & 132.3 & 129.8 & 85.9 \%\\ % detection rate on MPJPE: 100-14.29
%		LCR-Demo & 170.6$ & 266.8 & 100-14.4\%\\ % without normalization
		\LCR-H36M{} & 240.9 & 238.5 & 171.7 & 37.6 \% \\ %  without normalization100-62.6
%		LCR-H36M$^\star$ & 174.1 & 171.7 & 37.6 \% \\ %  without normalization100-62.6
%		LCR-H36M & 240.9 & 238.5 & 100-66.8\% \\
		\direct-$\bar{\mI}$ & 196.0 & 194.8 & 182.2 & 98.9 \% \\
%		\direct-${\mI}$ & 213.6 & 212.2 & 187.4 & 98.7 \% \\
		\hline
	\end{tabular}%
	%	\end{small}
}
	\caption{\small \textbf{Accuracy of two-person NSD on the boxing dataset}, \NEW{as average over all detected persons}. NMPJPE$^\star$ is a version of NMPJPE that accounts for LCR's different skeleton dimensions. It normalize predictions before error computation with the 17$\times$17 linear map that aligns prediction and GT in the least squares sense.}
	\label{fig:boxing_quantitative}
	\vspace{-3mm}
\end{figure}

%% file: tex/fig_threePersons.tex
% !TEX root = ../top.tex
% !TEX spellcheck = en-US

\begin{figure}
	\centering
	%\resizebox{\linewidth}{!}
	\includegraphics[width=1\linewidth]{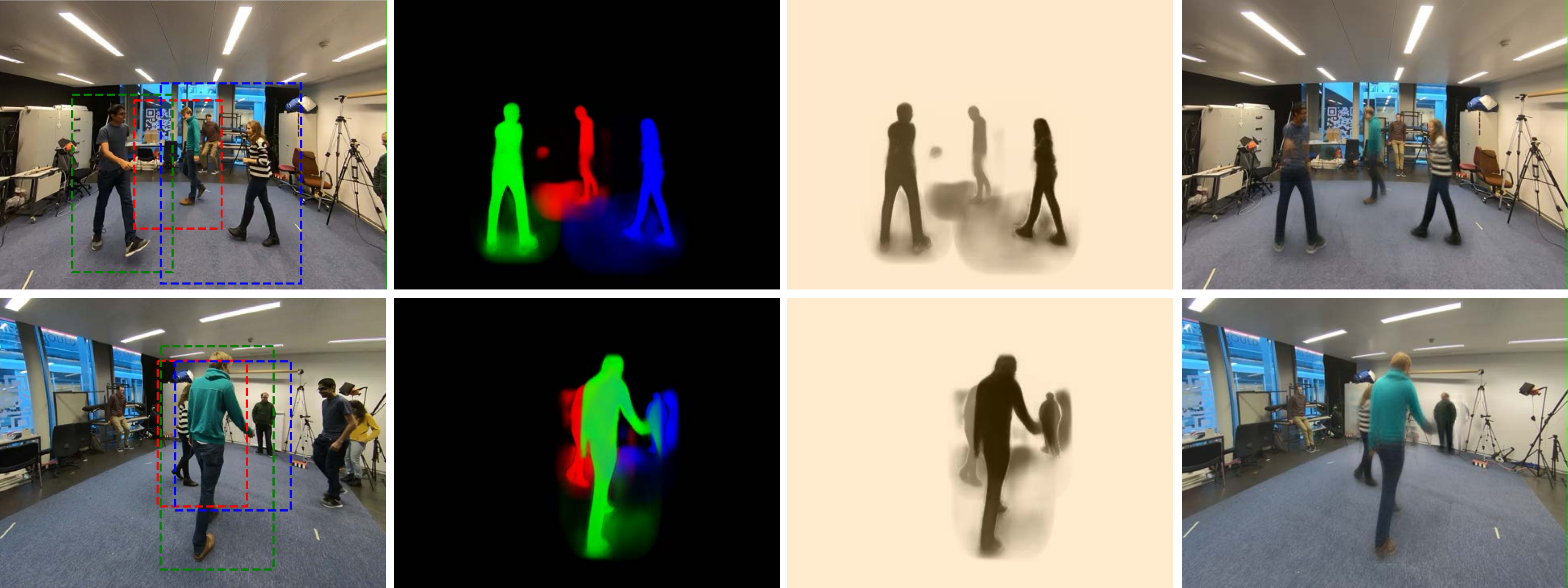}\\
	(a)\hspace{18mm}(b)\hspace{18mm}(c)\hspace{18mm}(d)
	\caption{\small \textbf{Three-person NSD training.} (a) The three detected bounding boxes. (b) Segmentation masks. (c) Depth ordering, where darker pixels are closer. (d) Re-synthesized images.}
	\label{fig:threePeople}
\end{figure}

%% file: tex/conc.tex
% !TEX root = ../top.tex
% !TEX spellcheck = en-US

\section{Conclusion}

We have proposed a \NEW{multi-view} self-supervision  approach to training a network to produce a hierarchical scene representation that is tailored for 3D human pose capture, yet general enough to be employed for other reconstruction tasks. It includes 3 levels of abstraction, \NEW{spatial layout} (bounding box and relative depth), instance segmentation (masks), and \NEW{body representation} (latent vectors that encode appearance and pose). Given that representation, very little annotated data suffices to train a secondary network to map it to a full 3D pose. The trained network can then operate without being given {\it a priori} locations for the people. It can compute their poses in parallel, even when they overlap.
%m in parallel with the pose even where there is more than one in the scene. 

In this work, we have limited ourselves to a few people in the scene. It serves well to our primary application domain of sports performance analysis, which demands high accuracy but where the number of athletes is known in advance. %Already this simplified case required several advances, such the bidirectional NVS.
In future work, we will extend this to a larger and unknown number of people. 

%Our evaluation on unknown number of subjects shows that the appearance matching layer is the weakest part in this direction and we plan to explore 

%because multiview-training databases featuring more than one person are rare. In fact, it 
% \PF{Say how and talk about how Pierre's work would fit in.}

%% file: tex/appendix.tex
% !TEX root = ../top.tex
% !TEX spellcheck = en-US

In this document, we supply additional implementation details, provide an ablation study, introduce focal spatial transformers, and explain the differentiable occlusion model in depth. %The supplemental HTML file contains additional qualitative results as video clips.

\section{Implementation Details}

\paragraph{Staged Training.}

Multi-person NSD requires to train $\cB$, $\cE$ and $\cD$ in staged fashion. First, we train all three networks without depth reasoning.
%In practice, we found that this stage, which trains all three networks, correctly localizes the subjects but inconsistently matches them across the views in the boxing scenario. Second, we re-initialize $\cE$ and $\cD$ to random values and incorporate the depth output of the detector. This leads to correct person associations.
Second, we re-initialize $\cE$ and $\cD$ to random values and incorporate the depth output of the detector.
In practice, we found that the first stage correctly localizes the subjects but inconsistently matches them across the views.
The second stage corrects the person associations. $\cP$ is trained in a third stage, keeping $\cB$ and $\cE$ fixed.
% but blurry reconstructions. 
% Third, we experimented with replacing the $L^2$ loss with an $L^1$ one. It required to keep $\cB$ with weights fixed from the previous stages, as we observed $\cB$ to diverge when using the $L^1$ loss.
% \MS{Do you have an intuition why?}\HR{It is known to be not so well behaved, but I'm not quite sure why it is so bad..}

\parag{Network architecture and hyperparameter.}
We use 18- and 50-layer residual networks for $\cB$ and $\cE$, respectively. They were pre-trained on ImageNet. For $\cD$ we employ a U-Net architecture~\cite{Ronneberger15} with 32, 64, 128, 256 feature channels in each stage. Following~\cite{Rhodin18b}, we define $\cP$ as a simple fully connected network with two layers of 1024 features and dropout with probability 0.5. All training stages are optimized for 200k iterations with Adam and a learning rate of 1e-3 for $\cB$ and $\cE$, and of 1e-4 for $\cD$ and $\cP$. We use an input image resolution of 910px$\times$512px and a batch size of 16 for the boxing dataset, 480px$\times$360px and a batch size of 12 for the three-person dataset, and 500px$\times$500px and a batch size of 32 for H36M. 
%self-supervised NSD and of 32 for 3D pose estimation. 
The loss $L(\cdot)$ in Eq.~1 of the main document
is implemented as the combination of a simple image loss on the pixel intensity and a perceptual loss on ResNet-ImageNet features. Both losses use the $L^2$ distance and the perceptual loss is weighted by a factor two.
$\cP$ is optimized with the mean squared error (MSE) loss.

The input images are whitened and the segmentation masks $\hat{S}$ are normalized to the range [0,1] before foreground-background blending.

\parag{Implementation.}
We use the PyTorch platform for NN training. To deal with the increased memory throughput due to using full-frame images instead of pre-processed crops, we use the NVVL loader~\cite{NVVL18}. It loads videos in compressed format and decodes them efficiently on the GPU.

\NEW{
\paragraph{Inverse spatial transformers}

The inverse spatial transformer maps from the small bounding box crop to full-frame. 
To handle regions without source pixels,
we use the Pytorch grid sample function with
padding. Zero padding is used for the segmentation and border padding
for the decoded image.
The resulting partially–filled but full-frame images are
completed by blending those regions where the segmentation mask is 0 with the background via Eq. 3 and Eq. 6. (main paper).
}

\NEW{
\section{Introducing Focal Spatial Transformers}

Spatial transformers \cite{Jaderberg15} make image transformations differentiable. We use them for localizing the subjects and noticed that their unsupervised training slows down convergence.
To counteract, we propose to encourage them to focus on the subject, which should be at the center of the crop. We therefore define a smooth mask $\mG$, which we model as a bump function $\mG = \exp\left(1 - \frac{1}{1 - \sqrt{x^4 + y^4}}\right)$ with $C^\infty$ smoothness on the compact crop window. We then
post-multiply the spatial transformer operation with $\mG$, which yields the focal spatial transformer (FST)
\begin{equation}
\tilde{\cT}(\vb,{\mI}) = \cT(\vb,{\mI}) \mG.
\end{equation}
Conversely, we use a pre-multiplication for the inverse unwarping operation, that is,  $\tilde{\cT}^{-1}(\vb,\bar{\mI}) = \ST^{-1}(\vb,\mG \bar{\mI})$. 
}

\section{Ablation Study}

\input{tex/fig_training_progress}

All of our model components contribute to the success of NSD. Not using weak priors leads to divergent training, as shown in the second row of Fig.~\ref{fig:training_progress}. Convergence is significantly slower without using focal spatial transformers, as shown in the third row of Fig.~\ref{fig:training_progress}. 
Fig.~\ref{fig:ablation} highlights the importance of depth handling and Bi-NVS. The top row shows that simple image encoding separates the subjects roughly, but leads to bleeding between the subject instances.
Using Bi-NVS without depth information creates two separate masks, but partial occlusions, in this example at the arms, are not resolved. By contrast, our full model produces clear separation together with depth maps.

\input{tex/fig_eval_box_ablation.tex}

Furthermore, we evaluate the influence of using predicted and ground truth (GT) bounding boxes on pose estimation.  Unfortunately, the popular Human80K subset of H36M provides only cropped images and the recent PoseTrack challenge only full-frame input without 2D pose or bounding box annotation. To nevertheless compare the algorithm on the same dataset, we use the unofficial protocol from \cite{Rhodin18a}. We train self-supervised on all five training subjects of H36M and supervised on Subject 1 only.
Training \ours{} with GT bounding boxes instead of the ones produced by $\cB$ leads to a reduction in pose estimation error from $145.3$ to $124.4$ N-MPJPE. Such a shift is expected, since the bounding box location provides additional albeit unrealistic cues at test time.
%For instance the aspect ratio of the crop is already indicative of the 
%The GT version is equivalent to \cite{Rhodin18a}, up to the background blending strategy and using all 17 instead of 16 joints. The $122.2$ N-MPJPE matches with the original numbers reported in \cite{Rhodin18a} and thereby permits an indirect comparison this method.
%, except for the different background blending strategies.

\section{Differentiable Occlusion Model}

Our goal is an occlusion model that is smooth and thereby differentiable in the depth ordering of objects. It should also be physically correct in that a partial occluder is as much visible as it reduces the visibility of the further occluded objects---the visibility of all objects must sum to one.

While occlusion and dis-occlusion of solid objects is generally non-differentiable,
these properties can be attained by smoothing the scene to be partially translucent and modeling physical light transport in a participating medium without scattering. In the following, we review the model used in \cite{Rhodin15,Rhodin16b} that approximates the scene with a set of Gaussian densities and explain our simplifications to it. 
%Under such assumptions, light travels straight and intensity  . 

In contrast to previous work that approximated arbitrary scenes consisting of multiple objects by hundreds of Gaussians, we assume that people are sufficiently separated and model each with a single depth plane, see Fig.~6 in the main document.
For the sake of smoothness, we make each plane partially translucent with a smooth Gaussian density in the $z$ direction.
To model the complex shape of humans, we consider different densities for each pixel. In practice, we use the generated segmentation masks to perform a kind of alpha blending.

We model how light travels in the $z$ direction along a view ray. In the following, we consider a single pixel and apply this model to each of the pixels, with varying opacity for each layer and pixel.
Beer-Lambert law states that the transmittance function from a point $s$ to the observer at $-\infty$ in a participating medium decays exponentially with the traversed density, that is,
\begin{align}
T(s) &= \exp\left(-\int_{-\infty}^{s} d(z) dz\right) \; ,
\end{align}
with $d(z)$ the density at point $z$ and assuming an orthographic projection with the observer at $-\infty$. Using a Gaussian density with means $(\mu_q)_q$ and $(\sigma_q)_q$ has the advantage that this integral can be written in closed form as
\begin{align}
\int_{-\infty}^{s} d(z) 
&= \int_{-\infty}^{s} \sum_q c_q G_q(z;\sigma_q,\mu_q) dz \nonumber\\
&= \int_{-\infty}^{s} \sum_q c_q \exp(-\frac{(z-\mu_q)^2}{2 \sigma_q^2}) dz \nonumber\\
%&= \sum_q \frac{\sigma_q c_q \sqrt{\pi}}{\sqrt{2}} \left(\erf(\frac{s-\mu_q}{\sqrt{2}\sigma_q})   -\erf(\frac{-\infty-\mu_q}{\sqrt{2}\sigma_q}) \right) \nonumber\\
&= \sum_q \frac{\sigma_q c_q \sqrt{\pi}}{\sqrt{2}} \left(\erf\left(\frac{s-\mu_q}{\sqrt{2}\sigma_q}\right) +1 \right) \; ,
\end{align}
in terms of the error function $\erf(s) = \frac{2}{\pi}\int_{0}^{s} \exp(-z^2) dz$.

In our case, we simplify this equation by assuming fixed Gaussian widths $\sigma=\frac{1}{\sqrt{2}}$ and magnitude 
$c_q = \frac{c_q' 2}{\sqrt{\pi}} $. This yields
\begin{align}
T(s) &=\exp\left(-\sum_q \frac{c_q \sqrt{\pi}}{2} \left(\erf(s-\mu_q) +1\right) \right)\nonumber\\
&= \exp\left(- \sum_q c_q' \left(\erf(s-\mu_q) + 1\right) \right) \;.
\end{align}
In this model, an object occludes as much as it is visible.
% which can be expressed through the negative derivative of transmittance, which is at each point split linearly into the contributions of the individual Gaussians
%Due to the exponential form,
The object's visibility $V(s)$ at position $s$ is proportional to the transmittance and the density of the object at $s$, that is,
\begin{align}
V_q(s) %&= -\frac{\partial T(s)}{\partial s} \nonumber\\
&= d(s)T(s) \; .
%&= \frac{c_q' 2}{\sqrt{\pi}} \exp(-(z-\mu_q)^2) \sum_q c_q' \left(\erf(s-\mu_q) + 1\right) 
\end{align}

For the background, which is assumed to be $\infty$ distant, we use the simplified model of \cite{Rhodin16b} expressed as
\begin{align}
T(\infty) &= \exp(-\sum_q c_q' \left(\erf(\infty-\mu_q) + 1 \right) ) \nonumber\\
&= \exp(-\sum_q c_q' \left(  2 \right) ) \; .
\end{align}
and the visibility of the background plane is equal to that remaining fraction $T(\infty)$.
The individual depth planes have been diffused in $z$ direction. To capture the entire visibility of one in relation to the other potentially intersecting depth layers, one has to integrate the point-wise visibility of the diffused density
\begin{align}
V_q = \int_{-\infty}^{\infty} V_q(s) ds \; .
\end{align}
This integral cannot be computed in terms of simple functions and was approximated by regular sampling in~\cite{Rhodin15}. Here we approximate it with a single sample at the Gaussian's position $\mu_q$, that is,
\begin{align}
V_q &\approx V_q(s) S_q \frac{2}{\sqrt{pi}}ds \nonumber\\
&\propto V_q(s) S_q \; .
\end{align}
The `lost' energy due to this drastic approximation can be inferred from $1-T(\infty)$, for which we have an analytic solution.
Assuming that the error is equally distributed across all Gaussians, we simply re-weight the visibility of each Gaussian by
\begin{align}
Z = \frac{1-\mT(\infty)}{\sum_j\mT(z_{v,j})\mS_{v',i}} \;,
\end{align}
so that their sum with the background visibility is exactly one. These simplifications ensure computational efficiency while maintaining smoothness and differentiability.

%% file: tex/fig_training_progress.tex
% !TEX root = ../supplemental.tex
% !TEX spellcheck = en-US

\begin{figure}
	\centering
	%\resizebox{\linewidth}{!}
	{
	\includegraphics[width=1\linewidth]{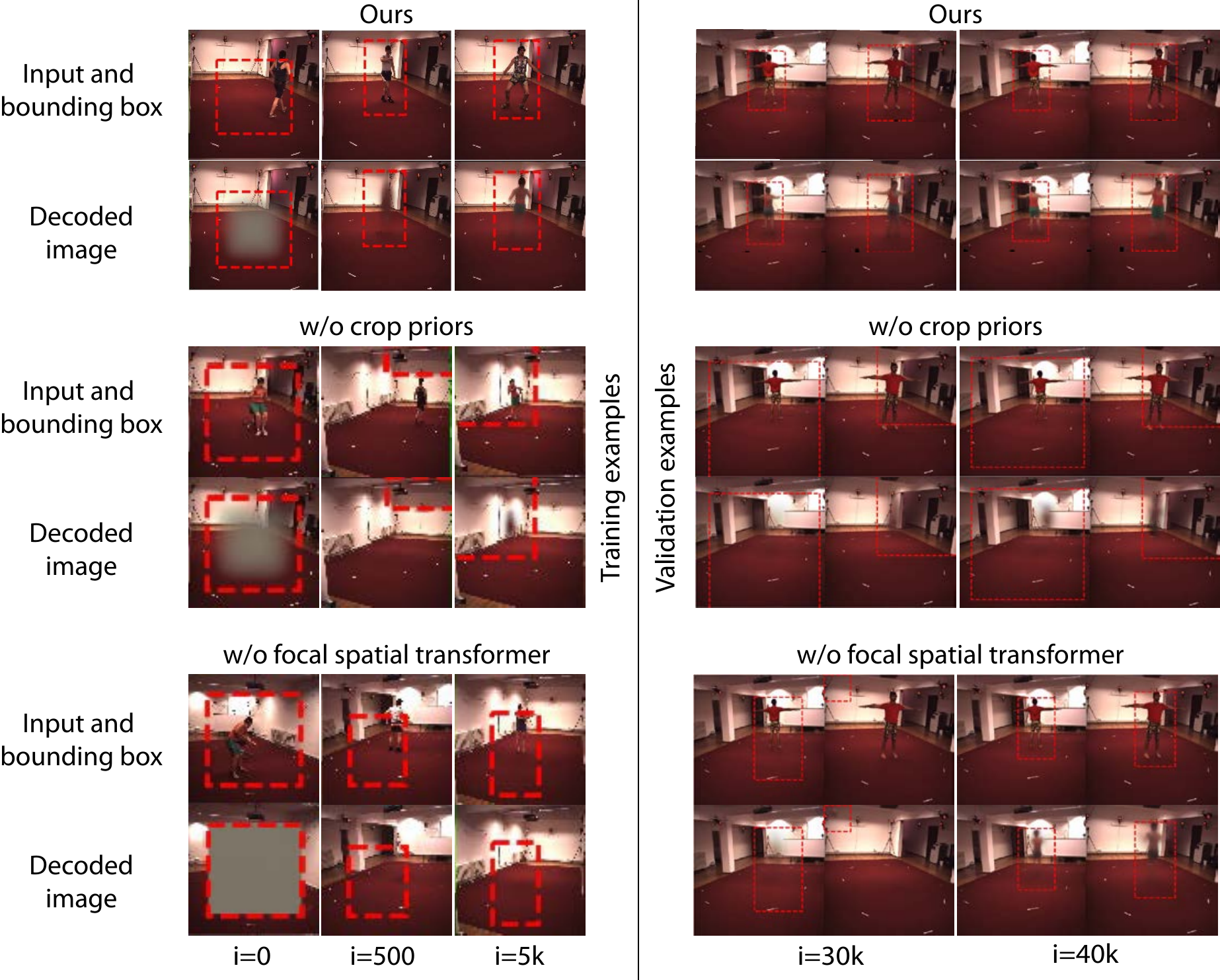}
}
	\caption{\textbf{Training progress} with respect to the number of performed gradient steps. Top row: With our full model the detector learns to localize the persons quickly (left) and Bi-NVS gives already reasonable reconstructions on the validation set after 30k iterations.
Central row: Without weak priors, training gets trapped in a local minima.
		Bottom row: With classical spatial transformers the detector converges much slower and, hence, also the Bi-NVS takes longer to converge and remains blurry after 40k iterations 
	}
	\label{fig:training_progress}
\end{figure}

%% file: tex/fig_eval_box_ablation.tex
% !TEX root = ../supplemental.tex
% !TEX spellcheck = en-US

\begin{figure}
	\centering
	%\resizebox{\linewidth}{!}
	{
	\includegraphics[width=1\linewidth]{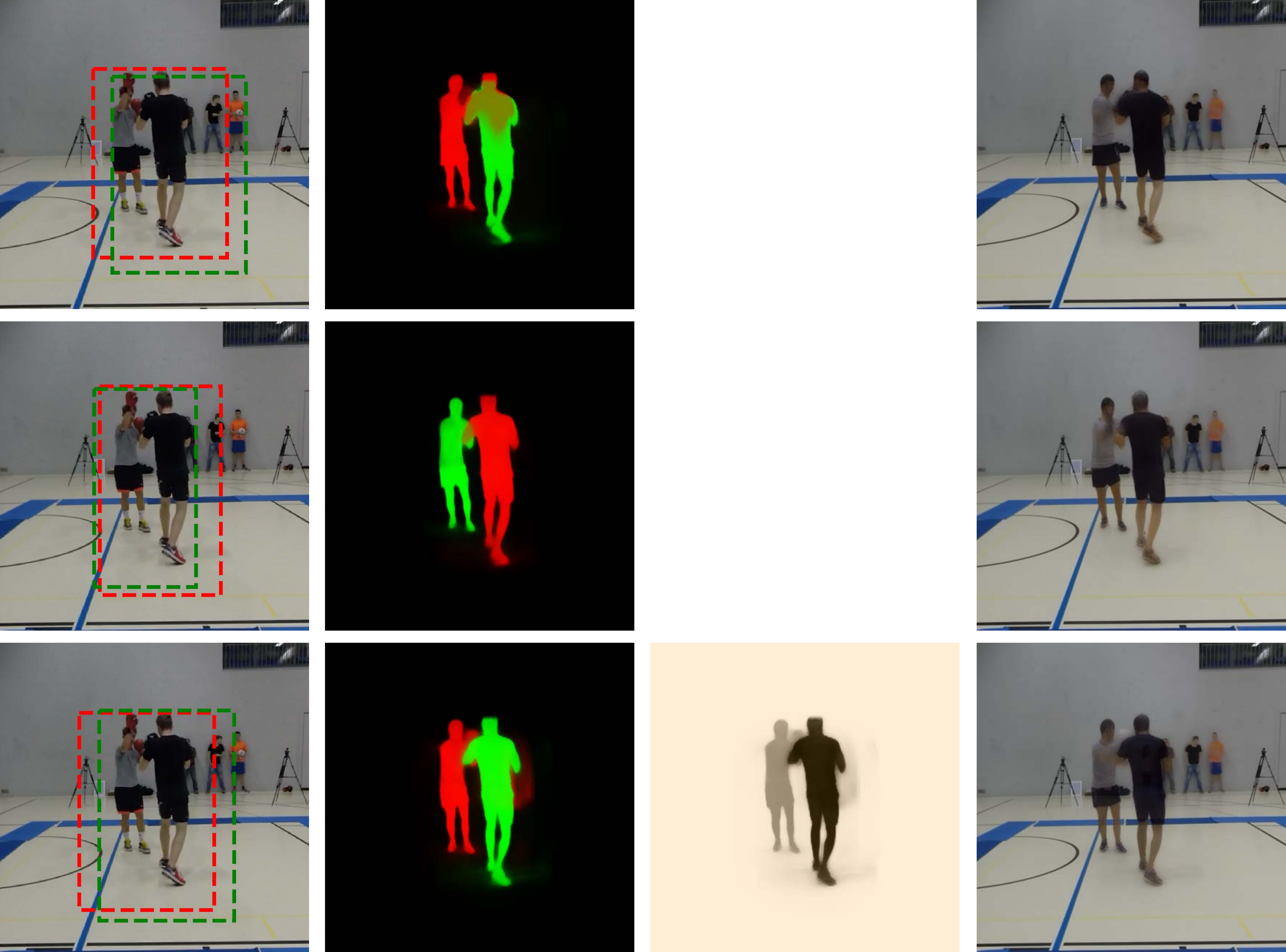}
}
	\caption{\textbf{Ablation study.} Top row: Without NVS task, that means using \auto, partially overlapping subjects merge.
		Central row: Without depth handling, overlapping parts blend in the segmentation mask. 
		Bottom row: NSD yields clear instance segmentation masks and depth map. 
	}
	\label{fig:ablation}
\end{figure}

%% file: top.bbl
\begin{thebibliography}{10}\itemsep=-1pt

\bibitem{Balakrishnan18}
G.~Balakrishnan, A.~Zhao, A.~Dalca, F.~Durand, and J.~Guttag.
\newblock {Synthesizing Images of Humans in Unseen Poses}.
\newblock In {\em Conference on Computer Vision and Pattern Recognition}, 2018.

\bibitem{Baque17b}
P.~Baqu{\'{e}}, F.~Fleuret, and P.~Fua.
\newblock {Deep Occlusion Reasoning for Multi-Camera Multi-Target Detection}.
\newblock In {\em International Conference on Computer Vision}, 2017.

\bibitem{Bas17}
A.~Bas, P.~Huber, W.~Smith, M.~Awais, and J.~Kittler.
\newblock {3D Morphable Models as Spatial Transformer Networks}.
\newblock {\em arXiv Preprint}, 2017.

\bibitem{Bengio12}
Y.~Bengio, A.~Courville, and P.~Vincent.
\newblock {Representation Learning: A Review and New Perspectives}.
\newblock {\em ArXiv e-prints}, 2012.

\bibitem{Burenius13}
M.~Burenius, J.~Sullivan, and S.~Carlsson.
\newblock {3D Pictorial Structures for Multiple View Articulated Pose
  Estimation}.
\newblock In {\em Conference on Computer Vision and Pattern Recognition}, 2013.

\bibitem{Captury}
{The Captury}.
\newblock http://thecaptury.com.

\bibitem{NVVL18}
J.~Casper, J.~Barker, and B.~Catanzaro.
\newblock Nvvl: Nvidia video loader.
\newblock \url{https://github.com/NVIDIA/nvvl}, 2018.

\bibitem{Chen16}
W.~Chen, H.~Wang, Y.~Li, H.~Su, Z.~Wang, C.~Tu, D.~Lischinski, D.~Cohen-or, and
  B.~Chen.
\newblock {Synthesizing Training Images for Boosting Human 3D Pose Estimation}.
\newblock In {\em 3DV}, 2016.

\bibitem{Chen16info}
X.~Chen, Y.~Duan, R.~Houthooft, J.~Schulman, I.~Sutskever, and P.~Abbeel.
\newblock {Infogan: Interpretable Representation Learning by Information
  Maximizing Generative Adversarial Nets}.
\newblock In {\em Advances in Neural Information Processing Systems}, pages
  2172--2180, 2016.

\bibitem{Cohen14}
T.~Cohen and M.~Welling.
\newblock {Transformation Properties of Learned Visual Representations}.
\newblock {\em arXiv Preprint}, 2014.

\bibitem{Dosovitskiy15}
A.~Dosovitskiy, J.~Springenberg, and T.~Brox.
\newblock {Learning to Generate Chairs with Convolutional Neural Networks}.
\newblock In {\em Conference on Computer Vision and Pattern Recognition}, 2015.

\bibitem{Dosovitskiy17}
A.~Dosovitskiy, J.~Springenberg, M.~Tatarchenko, and T.~Brox.
\newblock {Learning to Generate Chairs, Tables and Cars with Convolutional
  Networks}.
\newblock {\em IEEE Transactions on Pattern Analysis and Machine Intelligence},
  39(4):692--705, 2017.

\bibitem{Eslami18}
A.~Eslami, D.~Rezende, F.~Besse, F.~Viola, A.~Morcos, M.~Garnelo, A.~Ruderman,
  A.~Rusu, I.~Danihelka, K.~Gregor, D.~Reichert, L.~Buesing, T.~Weber,
  O.~Vinyals, D.~Rosenbaum, N.~Rabinowitz, H.~King, C.~Hillier, M.~Botvinick,
  D.~Wierstra, K.~Kavukcuoglu, and D.~Hassabis.
\newblock {Neural Scene Representation and Rendering}.
\newblock {\em Science}, 360(6394):1204--1210, 2018.

\bibitem{Flynn16}
J.~Flynn, I.~Neulander, J.~Philbin, and N.~Snavely.
\newblock {Deepstereo: Learning to Predict New Views from the World's Imagery}.
\newblock In {\em Conference on Computer Vision and Pattern Recognition}, pages
  5515--5524, 2016.

\bibitem{Gadelha16}
M.~Gadelha, S.~Maji, and R.~Wang.
\newblock {3D Shape Induction from 2D Views of Multiple Objects}.
\newblock {\em arXiv preprint arXiv:1612.05872}, 2016.

\bibitem{Garg16}
R.~Garg, G.~Carneiro, and I.~Reid.
\newblock {Unsupervised CNN for Single View Depth Estimation: Geometry to the
  Rescue}.
\newblock In {\em European Conference on Computer Vision}, pages 740--756,
  2016.

\bibitem{Grant16}
E.~Grant, P.~Kohli, and van M.~Gerven.
\newblock {Deep Disentangled Representations for Volumetric Reconstruction}.
\newblock In {\em European Conference on Computer Vision}, pages 266--279,
  2016.

\bibitem{Gunel18}
S.~G{\"{u}}nel, H.~Rhodin, and P.~Fua.
\newblock {What Face and Body Shapes Can Tell About Height}.
\newblock {\em arXiv}, 2018.

\bibitem{Higgins16}
I.~Higgins, L.~Matthey, A.~Pal, C.~Burgess, X.~Glorot, M.~Botvinick,
  S.~Mohamed, and A.~Lerchner.
\newblock {Beta-Vae: Learning Basic Visual Concepts with a Constrained
  Variational Framework}.
\newblock 2016.

\bibitem{Hinton11}
G.~Hinton, A.~Krizhevsky, and S.~Wang.
\newblock {Transforming Auto-Encoders}.
\newblock In {\em International Conference on Artificial Neural Networks},
  pages 44--51, 2011.

\bibitem{Ionescu14b}
C.~Ionescu, J.~Carreira, and C.~Sminchisescu.
\newblock {Iterated Second-Order Label Sensitive Pooling for 3D Human Pose
  Estimation}.
\newblock In {\em Conference on Computer Vision and Pattern Recognition}, 2014.

\bibitem{Jaderberg15}
M.~Jaderberg, K.~Simonyan, A.~Zisserman, and K.~Kavukcuoglu.
\newblock {Spatial Transformer Networks}.
\newblock In {\em Advances in Neural Information Processing Systems}, pages
  2017--2025, 2015.

\bibitem{Joo15}
H.~Joo, H.~Liu, L.~Tan, L.~Gui, B.~Nabbe, I.~Matthews, T.~Kanade, S.~Nobuhara,
  and Y.~Sheikh.
\newblock {Panoptic Studio: A Massively Multiview System for Social Motion
  Capture}.
\newblock In {\em International Conference on Computer Vision}, 2015.

\bibitem{Kanazawa18b}
A.~Kanazawa, M.~Black, D.~Jacobs, and J.~Malik.
\newblock {End-To-End Recovery of Human Shape and Pose}.
\newblock In {\em Conference on Computer Vision and Pattern Recognition}, 2018.

\bibitem{Kar17}
A.~Kar, C.~H{\"a}ne, and J.~Malik.
\newblock {Learning a Multi-View Stereo Machine}.
\newblock In {\em Advances in Neural Information Processing Systems}, pages
  364--375, 2017.

\bibitem{Kim18b}
H.~Kim, P.~Garrido, A.~Tewari, W.~Xu, J.~Thies, M.~Nie{\ss}ner, P.~P{\'e}rez,
  C.~Richardt, M.~Zollh{\"o}fer, and C.~Theobalt.
\newblock {Deep Video Portraits}.
\newblock {\em arXiv preprint arXiv:1805.11714}, 2018.

\bibitem{Kim17}
H.~Kim, M.~Zollh{\"o}fer, A.~Tewari, J.~Thies, C.~Richardt, and C.~Theobalt.
\newblock {Inversefacenet: Deep Single-Shot Inverse Face Rendering from a
  Single Image}.
\newblock {\em arXiv Preprint}, 2017.

\bibitem{Krizhevsky12}
A.~Krizhevsky, I.~Sutskever, and G.~Hinton.
\newblock {{ImageNet} Classification with Deep Convolutional Neural Networks}.
\newblock In {\em Advances in Neural Information Processing Systems}, pages
  1106--1114, 2012.

\bibitem{Kudo18}
Y.~Kudo, K.~Ogaki, Y.~Matsui, and Y.~Odagiri.
\newblock Unsupervised adversarial learning of 3d human pose from 2d joint
  locations.
\newblock {\em CoRR}, abs/1803.08244, 2018.

\bibitem{Kulkarni15}
T.~D. Kulkarni, W.~Whitney, P.~Kohli, and J.~B. Tenenbaum.
\newblock {Deep Convolutional Inverse Graphics Network}.
\newblock In {\em arXiv}, 2015.

\bibitem{Kundu18}
A.~Kundu, Y.~Li, and J.~Rehg.
\newblock 3d-rcnn: Instance-level 3d object reconstruction via
  render-and-compare.
\newblock In {\em Conference on Computer Vision and Pattern Recognition}, pages
  3559--3568, 2018.

\bibitem{Lassner17b}
C.~Lassner, G.~Pons-Moll, and P.~Gehler.
\newblock {A Generative Model of People in Clothing}.
\newblock {\em arXiv Preprint}, 2017.

\bibitem{Lassner17a}
C.~Lassner, J.~Romero, M.~Kiefel, F.~Bogo, M.~Black, and P.~Gehler.
\newblock {Unite the People: Closing the Loop Between 3D and 2D Human
  Representations}.
\newblock In {\em Conference on Computer Vision and Pattern Recognition}, 2017.

\bibitem{Ma17}
L.~Ma, X.~Jia, Q.~Sun, B.~Schiele, T.~Tuytelaars, and L.~V. Gool.
\newblock {Pose Guided Person Image Generation}.
\newblock In {\em Advances in Neural Information Processing Systems}, pages
  405--415, 2017.

\bibitem{Martinez17}
J.~Martinez, R.~Hossain, J.~Romero, and J.~Little.
\newblock {A Simple Yet Effective Baseline for 3D Human Pose Estimation}.
\newblock In {\em International Conference on Computer Vision}, 2017.

\bibitem{Mehta17a}
D.~Mehta, H.~Rhodin, D.~Casas, P.~Fua, O.~Sotnychenko, W.~Xu, and C.~Theobalt.
\newblock {Monocular 3D Human Pose Estimation in the Wild Using Improved CNN
  Supervision}.
\newblock In {\em International Conference on 3D Vision}, 2017.

\bibitem{Mehta18}
D.~Mehta, O.~Sotnychenko, F.~Mueller, W.~Xu, S.~Sridhar, G.~Pons-moll, and
  C.~Theobalt.
\newblock {Single-Shot Multi-Person 3D Pose Estimation from Monocular RGB}.
\newblock In {\em 3DV}, 2018.

\bibitem{Mehta17b}
D.~Mehta, S.~Sridhar, O.~Sotnychenko, H.~Rhodin, M.~Shafiei, H.~Seidel, W.~Xu,
  D.~Casas, and C.~Theobalt.
\newblock {Vnect: Real-Time 3D Human Pose Estimation with a Single RGB Camera}.
\newblock In {\em ACM SIGGRAPH}, 2017.

\bibitem{Moreno17}
F.~Moreno-noguer.
\newblock {3D Human Pose Estimation from a Single Image via Distance Matrix
  Regression}.
\newblock In {\em Conference on Computer Vision and Pattern Recognition}, 2017.

\bibitem{Park17}
E.~Park, J.~Yang, E.~Yumer, D.~Ceylan, and A.~Berg.
\newblock {Transformation-Grounded Image Generation Network for Novel 3D View
  Synthesis}.
\newblock In {\em Conference on Computer Vision and Pattern Recognition}, pages
  702--711, 2017.

\bibitem{Pavlakos18a}
G.~Pavlakos, X.~Zhou, and K.~Daniilidis.
\newblock {Ordinal Depth Supervision for 3D Human Pose Estimation}.
\newblock {\em Conference on Computer Vision and Pattern Recognition}, 2018.

\bibitem{Pavlakos16}
G.~Pavlakos, X.~Zhou, K.~Derpanis, G.~Konstantinos, and K.~Daniilidis.
\newblock {Coarse-To-Fine Volumetric Prediction for Single-Image 3D Human
  Pose}.
\newblock In {\em Conference on Computer Vision and Pattern Recognition}, 2017.

\bibitem{Pavlakos17}
G.~Pavlakos, X.~Zhou, K.~D.~G. Konstantinos, and D.~Kostas.
\newblock {Harvesting Multiple Views for Marker-Less 3D Human Pose
  Annotations}.
\newblock In {\em Conference on Computer Vision and Pattern Recognition}, 2017.

\bibitem{Pavlakos18}
G.~Pavlakos, L.~Zhu, X.~Zhou, and K.~Daniilidis.
\newblock {Learning to Estimate 3{D} Human Pose and Shape from a Single Color
  Image}.
\newblock In {\em Conference on Computer Vision and Pattern Recognition}, 2018.

\bibitem{PonsMoll14}
G.~Pons-Moll, D.~F. Fleet, and B.~Rosenhahn.
\newblock {Posebits for Monocular Human Pose Estimation}.
\newblock In {\em Conference on Computer Vision and Pattern Recognition}, 2014.

\bibitem{Popa17}
A.-I. Popa, M.~Zanfir, and C.~Sminchisescu.
\newblock {Deep Multitask Architecture for Integrated 2D and 3D Human Sensing}.
\newblock In {\em Conference on Computer Vision and Pattern Recognition}, 2017.

\bibitem{Rezende16}
D.~Rezende, S.~Eslami, S.~Mohamed, P.~Battaglia, M.~Jaderberg, and N.~Heess.
\newblock {Unsupervised Learning of 3D Structure from Images}.
\newblock In {\em Advances in Neural Information Processing Systems}, pages
  4996--5004, 2016.

\bibitem{Rhodin16b}
H.~Rhodin, N.~Robertini, D.~Casas, C.~Richardt, H.-P. Seidel, and C.~Theobalt.
\newblock {General Automatic Human Shape and Motion Capture Using Volumetric
  Contour Cues}.
\newblock In {\em European Conference on Computer Vision}, 2016.

\bibitem{Rhodin15}
H.~Rhodin, N.~Robertini, C.~Richardt, H.-P. Seidel, and C.~Theobalt.
\newblock {A Versatile Scene Model with Differentiable Visibility Applied to
  Generative Pose Estimation}.
\newblock In {\em International Conference on Computer Vision}, December 2015.

\bibitem{Rhodin18b}
H.~Rhodin, M.~Salzmann, and P.~Fua.
\newblock {Unsupervised Geometry-Aware Representation for 3D Human Pose
  Estimation}.
\newblock In {\em European Conference on Computer Vision}, 2018.

\bibitem{Rhodin18a}
H.~Rhodin, J.~Spoerri, I.~Katircioglu, V.~Constantin, F.~Meyer, E.~Moeller,
  M.~Salzmann, and P.~Fua.
\newblock {Learning Monocular 3D Human Pose Estimation from Multi-View Images}.
\newblock In {\em Conference on Computer Vision and Pattern Recognition}, 2018.

\bibitem{Rogez17}
G.~Rogez, P.~Weinzaepfel, and C.~Schmid.
\newblock {Lcr-Net: Localization-Classification-Regression for Human Pose}.
\newblock In {\em Conference on Computer Vision and Pattern Recognition}, 2017.

\bibitem{Rogez18}
G.~Rogez, P.~Weinzaepfel, and C.~Schmid.
\newblock {Lcr-Net++: Multi-Person 2D and 3D Pose Detection in Natural Images}.
\newblock In {\em arXiv preprint arXiv:1803.00455}, 2018.

\bibitem{Ronneberger15}
O.~Ronneberger, P.~Fischer, and T.~Brox.
\newblock {{U-Net}: Convolutional Networks for Biomedical Image Segmentation}.
\newblock In {\em Conference on Medical Image Computing and Computer Assisted
  Intervention}, pages 234--241, 2015.

\bibitem{Shu17a}
Z.~Shu, E.~Yumer, S.~Hadap, K.~Sunkavalli, E.~Shechtman, and D.~Samaras.
\newblock {Neural Face Editing with Intrinsic Image Disentangling}.
\newblock In {\em Conference on Computer Vision and Pattern Recognition}, 2017.

\bibitem{Si18}
C.~Si, W.~Wang, L.~Wang, and T.~Tan.
\newblock {Multistage Adversarial Losses for Pose-Based Human Image Synthesis}.
\newblock In {\em Conference on Computer Vision and Pattern Recognition}, June
  2018.

\bibitem{Sigal06}
L.~Sigal and M.~Black.
\newblock {Humaneva: Synchronized Video and Motion Capture Dataset for
  Evaluation of Articulated Human Motion}.
\newblock Technical report, Department of Computer Science, Brown University,
  2006.

\bibitem{Simon17}
T.~Simon, H.~Joo, I.~Matthews, and Y.~Sheikh.
\newblock {Hand Keypoint Detection in Single Images Using Multiview
  Bootstrapping}.
\newblock In {\em Conference on Computer Vision and Pattern Recognition}, 2017.

\bibitem{Simonyan15}
K.~Simonyan and A.~Zisserman.
\newblock {Very Deep Convolutional Networks for Large-Scale Image Recognition}.
\newblock In {\em International Conference on Learning Representations}, 2015.

\bibitem{Spurr18}
A.~Spurr, J.~Song, S.~Park, and O.~Hilliges.
\newblock {Cross-Modal Deep Variational Hand Pose Estimation}.
\newblock In {\em Conference on Computer Vision and Pattern Recognition}, pages
  89--98, 2018.

\bibitem{Sun17}
X.~Sun, J.~Shang, S.~Liang, and Y.~Wei.
\newblock {Compositional Human Pose Regression}.
\newblock In {\em International Conference on Computer Vision}, 2017.

\bibitem{Tan17}
J.~Tan, I.~Budvytis, and R.~Cipolla.
\newblock {Indirect Deep Structured Learning for 3D Human Body Shape and Pose
  Prediction}.
\newblock In {\em British Machine Vision Conference}, 2017.

\bibitem{Tatarchenko15}
M.~Tatarchenko, A.~Dosovitskiy, and T.~Brox.
\newblock {Single-View to Multi-View: Reconstructing Unseen Views with a
  Convolutional Network}.
\newblock {\em CoRR abs/1511.06702}, 1:2, 2015.

\bibitem{Tatarchenko16}
M.~Tatarchenko, A.~Dosovitskiy, and T.~Brox.
\newblock {Multi-View 3D Models from Single Images with a Convolutional
  Network}.
\newblock In {\em European Conference on Computer Vision}, pages 322--337,
  2016.

\bibitem{Tekin17a}
B.~Tekin, P.~Marquez-neila, M.~Salzmann, and P.~Fua.
\newblock {Learning to Fuse 2D and 3D Image Cues for Monocular Body Pose
  Estimation}.
\newblock In {\em International Conference on Computer Vision}, 2017.

\bibitem{Tewari17}
A.~Tewari, M.~Zollh{\"o}fer, H.~Kim, P.~Garrido, F.~Bernard, P.~P{\'e}rez, and
  C.~Theobalt.
\newblock {MoFA: Model-Based Deep Convolutional Face Autoencoder for
  Unsupervised Monocular Reconstruction}.
\newblock {\em International Conference on Computer Vision}, 2017.

\bibitem{Thewlis17b}
J.~Thewlis, H.~Bilen, and A.~Vedaldi.
\newblock {Unsupervised Learning of Object Frames by Dense Equivariant Image
  Labelling}.
\newblock In {\em Advances in Neural Information Processing Systems}, pages
  844--855, 2017.

\bibitem{Thewlis17a}
J.~Thewlis, H.~Bilen, and A.~Vedaldi.
\newblock {Unsupervised Learning of Object Landmarks by Factorized Spatial
  Embeddings}.
\newblock In {\em International Conference on Computer Vision}, 2017.

\bibitem{Tome17}
D.~Tome, C.~Russell, and L.~Agapito.
\newblock {Lifting from the Deep: Convolutional 3D Pose Estimation from a
  Single Image}.
\newblock In {\em arXiv preprint, arXiv:1701.00295}, 2017.

\bibitem{Tran17}
L.~Tran, X.~Yin, and X.~Liu.
\newblock {Disentangled Representation Learning Gan for Pose-Invariant Face
  Recognition}.
\newblock In {\em Conference on Computer Vision and Pattern Recognition},
  page~7, 2017.

\bibitem{Tulsiani18}
S.~Tulsiani, A.~Efros, and J.~Malik.
\newblock {Multi-View Consistency as Supervisory Signal for Learning Shape and
  Pose Prediction}.
\newblock {\em arXiv Preprint}, 2018.

\bibitem{Tulsiani18b}
S.~Tulsiani, S.~Gupta, D.~Fouhey, A.~Efros, and J.~Malik.
\newblock Factoring shape, pose, and layout from the 2d image of a 3d scene.
\newblock In {\em CVPR}, pages 302--310, 2018.

\bibitem{Tulsiani17}
S.~Tulsiani, T.~Zhou, A.~Efros, and J.~Malik.
\newblock {Multi-View Supervision for Single-View Reconstruction via
  Differentiable Ray Consistency}.
\newblock In {\em Conference on Computer Vision and Pattern Recognition},
  page~3, 2017.

\bibitem{Tung17}
H.-Y. Tung, A.~Harley, W.~Seto, and K.~Fragkiadaki.
\newblock {Adversarial Inverse Graphics Networks: Learning 2D-To-3D Lifting and
  Image-To-Image Translation from Unpaired Supervision}.
\newblock In {\em International Conference on Computer Vision}, 2017.

\bibitem{Tung17self}
H.-Y. Tung, H.-W. Tung, E.~Yumer, and K.~Fragkiadaki.
\newblock {Self-Supervised Learning of Motion Capture}.
\newblock In {\em Advances in Neural Information Processing Systems}, pages
  5242--5252, 2017.

\bibitem{Varol18}
G.~Varol, D.~Ceylan, B.~Russell, J.~Yang, E.~Yumer, I.~Laptev, and C.~Schmid.
\newblock {{BodyNet}: Volumetric Inference of {3D} Human Body Shapes}.
\newblock In {\em European Conference on Computer Vision}, 2018.

\bibitem{Varol17}
G.~Varol, J.~Romero, X.~Martin, N.~Mahmood, M.~Black, I.~Laptev, and C.~Schmid.
\newblock {Learning from Synthetic Humans}.
\newblock In {\em Conference on Computer Vision and Pattern Recognition}, 2017.

\bibitem{Marcard18}
T.~von Marcard, R.~Henschel, M.~Black, B.~Rosenhahn, and G.~Pons-Moll.
\newblock {Recovering Accurate 3D Human Pose in the Wild Using IMUs and a
  Moving Camera}.
\newblock In {\em European Conference on Computer Vision}, 2018.

\bibitem{Wang18d}
T.-C. Wang, M.-Y. Liu, J.-Y. Zhu, G.~Liu, A.~Tao, J.~Kautz, and B.~Catanzaro.
\newblock {Video-To-Video Synthesis}.
\newblock In {\em Advances in Neural Information Processing Systems}, 2018.

\bibitem{Worrall17}
D.~Worrall, S.~Garbin, D.~Turmukhambetov, and G.~Brostow.
\newblock {Interpretable Transformations with Encoder-Decoder Networks}.
\newblock In {\em International Conference on Computer Vision}, 2017.

\bibitem{Yan16}
X.~Yan, J.~Yang, E.~Yumer, Y.~Guo, and H.~Lee.
\newblock {Perspective Transformer Nets: Learning Single-View 3D Object
  Reconstruction Without 3D Supervision}.
\newblock In {\em Advances in Neural Information Processing Systems}, pages
  1696--1704, 2016.

\bibitem{Yang15weakly}
J.~Yang, S.~Reed, M.-H. Yang, and H.~Lee.
\newblock {Weakly-Supervised Disentangling with Recurrent Transformations for
  3D View Synthesis}.
\newblock In {\em Advances in Neural Information Processing Systems}, pages
  1099--1107, 2015.

\bibitem{Yang18b}
W.~Yang, W.~Ouyang, X.~Wang, J.~Ren, H.~Li, and X.~Wang.
\newblock {3D Human Pose Estimation in the Wild by Adversarial Learning}.
\newblock {\em Conference on Computer Vision and Pattern Recognition}, 2018.

\bibitem{Zamir16}
A.~R. Zamir, T.~Wekel, P.~Agrawal, J.~Malik, and S.~Savarese.
\newblock {Generic 3D Representation via Pose Estimation and Matching}.
\newblock In {\em European Conference on Computer Vision}, 2016.

\bibitem{Zanfir18a}
A.~Zanfir, E.~Marinoiu, and C.~Sminchisescu.
\newblock {Monocular 3D Pose and Shape Estimation of Multiple People in Natural
  Scenes - the Importance of Multiple Scene Constraints}.
\newblock In {\em Conference on Computer Vision and Pattern Recognition}, June
  2018.

\bibitem{Zanfir18b}
M.~Zanfir, A.-I. Popa, A.~Zanfir, and C.~Sminchisescu.
\newblock {Human Appearance Transfer}.
\newblock In {\em Conference on Computer Vision and Pattern Recognition}, pages
  5391--5399, 2018.

\bibitem{Zhang18b}
Y.~Zhang, Y.~Guo, Y.~Jin, Y.~Luo, Z.~He, and H.~Lee.
\newblock {Unsupervised Discovery of Object Landmarks as Structural
  Representations}.
\newblock In {\em Conference on Computer Vision and Pattern Recognition}, pages
  2694--2703, 2018.

\bibitem{Zhao17}
B.~Zhao, X.~Wu, Z.-Q. Cheng, H.~Liu, and J.~Feng.
\newblock {Multi-View Image Generation from a Single-View}.
\newblock {\em arXiv preprint arXiv:1704.04886}, 2017.

\bibitem{Zhou17a}
T.~Zhou, M.~Brown, N.~Snavely, and D.~Lowe.
\newblock {Unsupervised Learning of Depth and Ego-Motion from Video}.
\newblock In {\em Conference on Computer Vision and Pattern Recognition}, 2017.

\bibitem{Zhou16c}
T.~Zhou, S.~Tulsiani, W.~Sun, J.~Malik, and A.~Efros.
\newblock {View Synthesis by Appearance Flow}.
\newblock In {\em European Conference on Computer Vision}, pages 286--301,
  2016.

\bibitem{Zhou17d}
X.~Zhou, Q.~Huang, X.~Sun, X.~Xue, and Y.~We.
\newblock {Weakly-Supervised Transfer for 3D Human Pose Estimation in the
  Wild}.
\newblock {\em arXiv Preprint}, 2017.

\bibitem{Zhu18c}
H.~Zhu, H.~Su, P.~Wang, X.~Cao, and R.~Yang.
\newblock View extrapolation of human body from a single image.
\newblock In {\em Conference on Computer Vision and Pattern Recognition}, pages
  4450--4459, 2018.

\bibitem{Zhu17c}
R.~Zhu, H.~Galoogahi, C.~Wang, and S.~Lucey.
\newblock Rethinking reprojection: Closing the loop for pose-aware shape
  reconstruction from a single image.
\newblock {\em International Conference on Computer Vision}, pages 57--65,
  2017.

\end{thebibliography}
